\documentclass{article}

\PassOptionsToPackage{numbers, compress}{natbib}
 \usepackage[preprint]{neurips_2026}

\usepackage[utf8]{inputenc} % allow utf-8 input
\usepackage[T1]{fontenc}    % use 8-bit T1 fonts
\usepackage{hyperref}       % hyperlinks
\usepackage{url}            % simple URL typesetting
\usepackage{booktabs}       % professional-quality tables
\usepackage{amsfonts}       % blackboard math symbols
\usepackage{nicefrac}       % compact symbols for 1/2, etc.
\usepackage{microtype}      % microtypography
\usepackage{xcolor}         % colors
\usepackage{graphicx}       % includegraphics
\usepackage{subcaption}     % subfigures
\usepackage{multirow}       % multirow tables
\usepackage{enumitem}       % compact lists
\usepackage{wrapfig}        % wrap figures
\usepackage{amsmath}

\usepackage{fontawesome}
\usepackage{float}
\usepackage{placeins}
\usepackage{longtable}
\usepackage{xspace}
\usepackage[capitalize]{cleveref}
\usepackage{todonotes}
\crefname{figure}{Fig.}{Figs.}
\Crefname{figure}{Fig.}{Figs.}
\crefname{section}{Sec.}{Secs.}
\Crefname{section}{Sec.}{Secs.}
\newcommand{\method}{\textsc{Agentic-imodels}\xspace}

\title{\method:\\Evolving agentic interpretability tools via autoresearch}

\author{%
  Chandan Singh$^1$\\
  \And
  Yan Shuo Tan$^2$\\
  \And
  Weijia Xu$^1$ \\
  \And
  Zelalem Gero$^1$ \\
  \AND
  Weiwei Yang$^1$ \\
  \And
  Michel Galley$^1$ \\
  \And
  Jianfeng Gao$^1$ \\
  \AND
   \\
  $^1$ Microsoft Research\\
  $^2$ National University of Singapore\\
}

\begin{document}

\maketitle

\begin{abstract}
Agentic data science (ADS) systems are rapidly improving their capability to autonomously analyze, fit, and interpret data,
potentially moving towards a future where agents conduct the vast majority of data-science work.
However, current ADS systems use statistical tools designed to be interpretable by humans, rather than interpretable by agents.
To address this, we introduce \method, an agentic autoresearch loop that evolves data-science tools designed to be interpretable by agents.
Specifically, it develops a library of \texttt{scikit-learn}-compatible regressors for tabular data that are optimized for both predictive performance and a novel LLM-based interpretability metric.
The metric measures a suite of LLM-graded tests that probe whether a fitted model's string representation is ``simulatable'' by an LLM, i.e. whether the LLM can answer questions about the model's behavior by reading its string output alone.
We find that the evolved models jointly improve predictive performance and agent-facing interpretability, generalizing to new datasets and new interpretability tests.
Furthermore, these evolved models improve downstream end-to-end ADS, increasing performance for Copilot CLI, Claude Code, and Codex on the BLADE benchmark by up to 73\%.\footnote{All code for reproducing and adapting \method is made available at \href{https://github.com/csinva/agentic-imodels}{\faGithub~github.com/csinva/agentic-imodels}. We additionally release the final library of \method we develop as a Python package / agent skill for easy integration into ADS workflows.}
\end{abstract}

\section{Introduction}
The data-science lifecycle has historically been a human-centered process end-to-end: a human defines the question, carries out the analysis using interpretable tools, and communicates the findings.
As agentic data science (ADS) systems grow more capable,
% the framing and interpretation of results remain human-centered, but the
intermediate analytical steps are increasingly delegated to agents, e.g. selecting models, interpreting coefficients, and drawing conclusions~\citep{guo2024ds,nie2026dsgym,chen2024scienceagentbench}.
This improvement in capabilities may be critical in accelerating scientific discovery~\citep{wang2023scientific,lu2024ai}, and in tackling emerging issues such as scalable oversight of AI systems~\citep{bowman2022measuring,gyevnar2025ai}.
However, the performance of ADS systems is limited by their tools for working with and interpreting data:
existing interpretability tools/models and their accompanying implementations produce outputs designed to be read by humans, often containing visualizations or intervenable components that may be difficult for agents to parse~\citep{rudin2018please,murdoch2019Definitions,breiman1984classification,tibshirani1996regression,singh2021imodels,nori2019interpretml}.
This mismatch can derail ADS systems, leading to unreliable analysis~\citep{asher2026sycophancy,rewolinski2026sanitychecksagenticdata} and obscured analytical choices~\citep{luo2025more,sivaraman2026meansendsupportingreasoning}.

This emerging problem leads us to rethink interpretability in the modern era: rather than designing data-science tools to be interpretable by humans, we aim to design tools that are interpretable \emph{by agents}.
% In response to this growing problem, we propose to build data-science tools that are interpretable \emph{to agents}.
Doing this requires adapting the rich literature on human-centered interpretable machine learning~\citep{kaur2020interpreting,hong2020human,doshi2017roadmap}, which has yielded interpretable models such as decision trees, generalized additive models, and rule lists~\citep{rudin2018please}.
This field has traditionally quantified interpretability in terms of concepts such as simulatability, sparsity, and modularity~\citep{lipton2016mythos,rudin2018please,murdoch2019Definitions},
ideally followed by quantitative evaluation with human experiments~\citep{lage2019evaluation,doshi2017roadmap}.
In our setting, we replace these human experiments with tests that measure whether an LLM can accurately simulate the model's behavior in terms of predictions, feature effects, and counterfactuals solely by reading its string representation.
These LLM-based tests provide a key practical advantage: they enable computing an \textit{agent interpretability score} for any model without the need for human experiments.

\begin{figure}[t]
    \centering
    \vspace{-10pt}
    \begin{minipage}[b]{0.31\textwidth}
        \includegraphics[width=\textwidth]{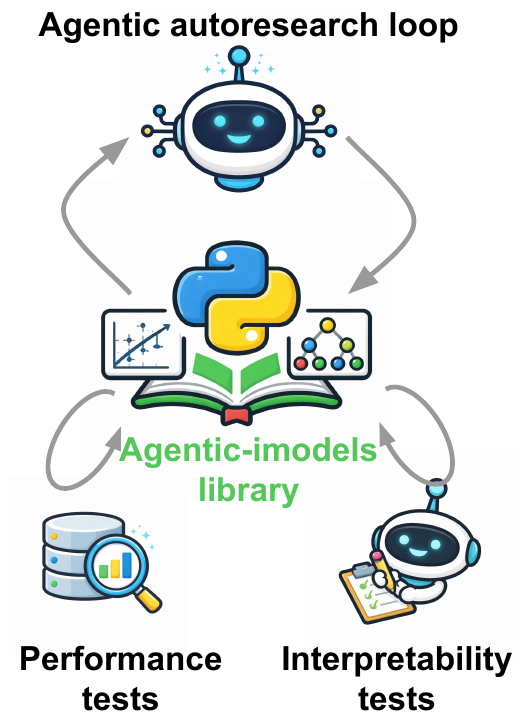}%
        \put(-115,180){\textbf{a}}%
    \end{minipage}%
    \hfill
    \begin{minipage}[b]{0.685\textwidth}
        \includegraphics[width=\textwidth]{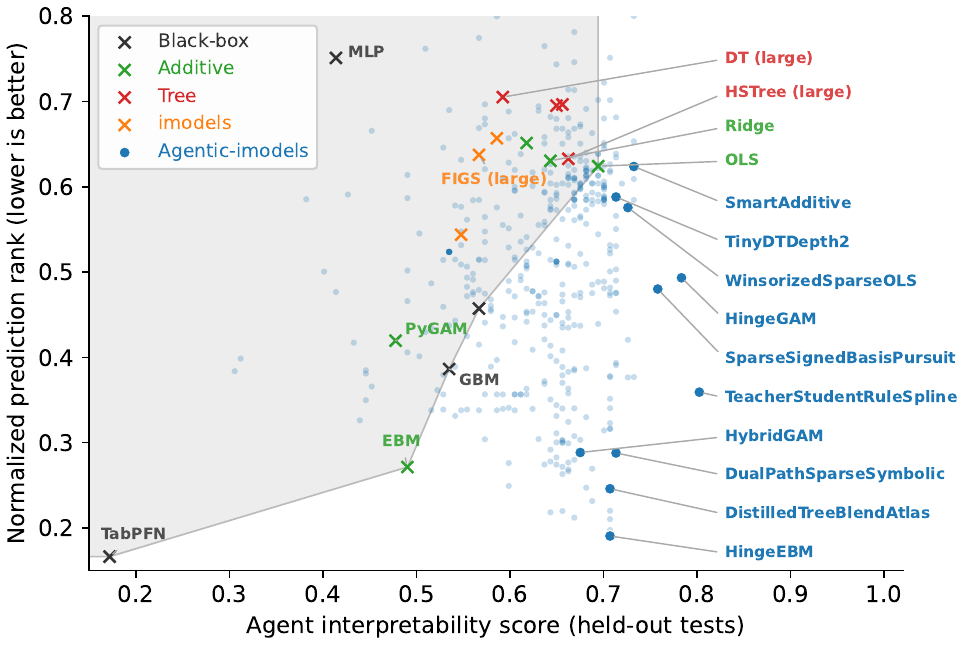}%
        \put(-275,180){\textbf{b}}%
    \end{minipage}
    \caption{(a) Overview of the \method autoresearch loop, which optimizes a Python class for predictive performance and agent interpretability (evaluated through LLM-based simulatability tests).
    (b) The discovered \method (blue points) improve the Pareto frontier of predictive performance and interpretability over baselines from the literature.
    See evaluation details in \cref{sec:main_results}.}
    \label{fig:overview}
\end{figure}

We use this agent interpretability score to guide the automated design of new interpretable models.
To do so, we propose \method, an autoresearch loop that prompts a coding agent (e.g. Claude Code) to iteratively modify a Python class so it simultaneously achieves a high interpretability score and strong predictive performance (\cref{fig:overview}a).
The loop consists of the coding agent generating candidate model modifications,
which are then evaluated using the interpretability score and predictive performance metrics across a large suite of tabular datasets.
The coding agent uses these metrics to refine the model, generating new candidate modifications and repeating the evaluation process until the model achieves a user-specified balance of predictive performance and interpretability.

In our experiments, we use \method to build \texttt{scikit-learn}-compatible regressors for tabular data.
We run the loop on a suite of 65 tabular datasets with both Claude Code and Codex, and find that it discovers new model classes that push the interpretability-performance frontier beyond a wide range of existing baselines (\cref{fig:overview}b), even for held-out datasets and new interpretability tests (\cref{sec:main_results}).
They are also useful downstream: equipping four different ADS agents with the evolved models improves their performance on the BLADE end-to-end benchmark~\citep{gu2024blade} by 8\%--73\% over standard interpretability tools (\cref{sec:e2e_results}).
Qualitatively, we find that the evolved \method combine existing algorithms in novel ways and present themselves in forms amenable to agent interpretation (\cref{sec:qualitative_model_descriptions}).

\section{Related work}
\label{sec:related}

\paragraph{Interpretable machine learning.}
Interpretable models, those whose reasoning can be directly understood by humans, have long been advocated as preferable to post-hoc explanations of black-box models~\citep{rudin2018please,rudin2021interpretable,murdoch2019Definitions}.
Prominent model families include decision trees~\citep{breiman1984classification,quinlan1986induction}, rule lists~\citep{letham2015interpretable}, generalized additive models (GAMs)~\citep{hastie1986generalized,lou2013accurate,caruana2015intelligible}, and sparse linear models~\citep{ustun2016supersparse}, implemented in popular packages such as \texttt{imodels}~\citep{singh2021imodels} and \texttt{interpretML}~\citep{nori2019interpretml}.
More recent approaches use LLMs to build interpretable models, e.g. Aug-imodels hand-designs methods to build an interpretable model from an LLM~\citep{singh2023augmenting}, and alternative works train transformers to directly produce an interpretable model, e.g.~\citep{mueller2024gamformer,zhuang2024learning}.
Our work extends this ecosystem by using agents to discover new agent-interpretable model classes.

\paragraph{Agentic data science (ADS).}
A rapidly growing literature studies how well coding and reasoning agents can conduct end-to-end data analysis, primarily by measuring agent capability on new benchmarks~\citep{nie2026dsgym,chen2024scienceagentbench,li2025ida,song2025evaluating,li2025autosdt,guo2024ds}, and by documenting ADS failure modes such as LLM-assisted $p$-hacking~\citep{asher2026sycophancy}, missing sanity checks~\citep{rewolinski2026sanitychecksagenticdata,sivaraman2026meansendsupportingreasoning,nam2025ds}, and unreliable LLM-driven annotation of data and findings~\citep{baumann2025large,tseng2025evaluating,luo2025more}.
A parallel thread design methods that use agents to directly describe data~\citep{zhong2023goaldd,babbar2025different,zhu2022gsclip,menon2023mantle} or describes features of a pre-trained foundation model~\citep{bills2023language,singh2023explainingmodules,han2026sage,shaham2024multimodal}.
In both threads the agent relies upon human-designed interpretability methods.
In contrast, we treat the coding agent as a designer of those methods, by having it directly optimize a model for LLM-graded interpretability.

\paragraph{Automated model discovery and autoresearch.}
Agents that generate and refine code for scientific purposes are a growing area of interest~\citep{lu2024ai,schmidgall2025agent,swanson2025virtual,yang2024large},
particularly agentic ``autoresearch'' loops against a verifiable outcome, e.g. AlphaEvolve~\citep{novikov2025alphaevolve} and related works~\citep{romera2023mathematical,singh2023explaining,wang2025thetaevolve,yuksekgonul2026learning}.
A related line of work focuses on reusable evolving text prompts (i.e. ``skills'') that can be used to improve agent performance, e.g. evolving skills collectively across users~\citep{ma2026skillclaw},
using multi-agent systems~\citep{zhou2026memento,alzubi2026evoskill},
or from scientific resources~\citep{shen2026skillfoundry}.
Alternative works more broadly seek to optimize agent harnesses end-to-end~\citep{lee2026meta,sengupta2026harbor}.
Recent test-time adaptation methods also improve agent capabilities through iterative self-refinement, potentially creating reusable skills as a byproduct~\citep{suzgun2026dynamic,wei2025evo,zhang2025darwin,zhuang2026test}.
Our work applies the autoresearch approach specifically to discovering ML models that are interpretable to agents.

\section{Methods}
\label{sec:methods}

The goal of \method is to automatically discover model classes that are both accurate and interpretable to LLMs.
It does this using an agentic autoresearch loop that simultaneously optimizes predictive performance and a novel interpretability metric, using a coding agent that iteratively proposes and refines model implementations (\cref{fig:overview}a).

In our instantiation of \method, each model class is a \texttt{scikit-learn}-compatible~\citep{pedregosa2011scikit} Python class with \texttt{fit}, \texttt{predict}, and \texttt{\_\_str\_\_}.
Predictive performance is measured by taking the average regression performance across a suite of tabular regression datasets.
Interpretability is measured by LLM-graded tests that probe whether an LLM can answer quantitative questions about the fitted model by reading only its \texttt{\_\_str\_\_} output.
The coding agent (e.g. Claude Code) is prompted to iteratively modify the model class, evaluate both metrics, and refine based on feedback.
We describe our choices for each component in detail below, but note that the \method loop can be generalized to optimize a broader set of interpretability tools.

\subsection{Evaluating predictive performance}
\label{sec:perf_eval}

Each model's predictive performance is measured using a suite of regression datasets.
For each dataset, each model is evaluated by fitting the model to a training split (and potentially tuning hyperparameters via cross-validation splits on the training set), and then computing the test performance on the dataset.
In our case we compute the test-set root mean squared error (RMSE) for each dataset in the suite.
To avoid sensitivity to noisy datasets, we rank models on each dataset and then average the ranks across datasets (rather than averaging RMSE directly).

\subsection{Evaluating agent interpretability via LLM-graded interpretability tests}
\label{sec:interp_tests}

For the \method autoresearch loop, we require a metric that serves as a proxy for the extent to which a model’s representation enables a downstream agent to reliably reason about its behavior.
We additionally desire the metric to be 
(1) automated, requiring no human evaluation and (2) model-agnostic, unlike existing proxies such as tree depth or number of coefficients which do not generalize across model families~\citep{murdoch2019Definitions,doshi2017roadmap,kaur2020interpreting}.
We simultaneously satisfy both criteria by defining the \textbf{Agent interpretability score} of a fitted model as the pass rate of LLM-based interpretability tests,
where each test evaluates whether an LLM can correctly answer a quantitative question about the model's predictions, given only the model's \texttt{\_\_str\_\_} output (note that it is impossible to score highly on the tests without the model's \texttt{\_\_str\_\_} providing the necessary information).

\begin{figure}[t]
    \centering
    \vspace{-20pt}
    \includegraphics[width=\textwidth]{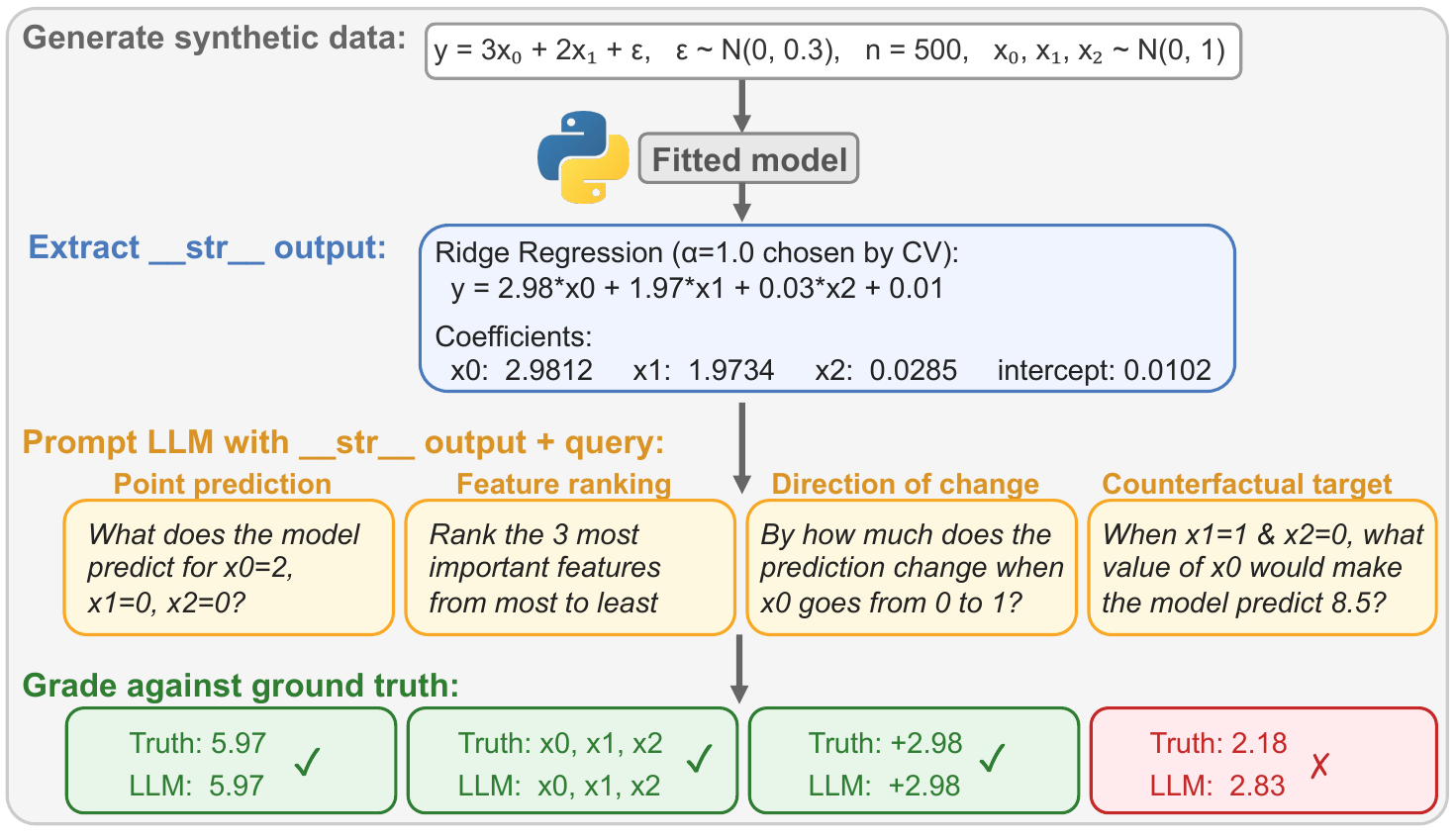}
    \caption{The interpretability test protocol, illustrated on Ridge regression with four of the 43 tests. Synthetic data is generated and the model is fit. The LLM receives only the \texttt{\_\_str\_\_} output and a query. The response is graded against the ground truth, which is obtained either from evaluating the fitted function or from knowledge of the ground-truth function.
    In some cases, the question itself requires evaluating the prediction function, e.g. in the \textit{counterfactual target} test the value \textit{8.50} is obtained by evaluating the model when setting $x_0 = 2.18, x_1 = 0.0, x_2 = 0.0$.
    Here the LLM passes three tests but fails the \textit{counterfactual target} test.
    Even though the information to compute the counterfactual target is present in the \texttt{\_\_str\_\_} output, the representation does not make it easy for the LLM to solve the inverse problem compared to other model representations (e.g. a decision tree that makes the predicted value easily apparent in leaf nodes).
    }
    \label{fig:hinge_example}
\end{figure}

Every test is broken down into a few steps, illustrated in \cref{fig:hinge_example} for a Ridge regression model.
First, synthetic data is generated from a known ground-truth function, here $y = 3x_0 + 2x_1 + \varepsilon$ with an irrelevant third feature $x_2$.
Second, the model is fit to this data, producing a fitted model whose \texttt{\_\_str\_\_} method returns a human-readable summary (in this case, the coefficient equation \texttt{y = 2.98*x0 + 1.97*x1 + 0.03*x2 + 0.01}, along with some extra printed information such as the regularization parameter and cross-validation score).
Third, an LLM receives only this string and a quantitative query about the model, e.g. ``What does this model predict for $x_0{=}2, x_1{=}0, x_2{=}0$?''.
Each response is then graded against the ground truth (with numerical tolerance), yielding a binary pass/fail per test.
For simplicity, \cref{fig:hinge_example} shows four tests using the same synthetic data / model, but in actuality, the tests span different synthetic datasets / models to cover diverse scenarios.

Drawing from prior work on evaluating interpretability via human experiments~\citep{doshi2017roadmap,lage2019evaluation,kaur2020interpreting}, we develop a total of 200 tests grouped into six categories:
\begin{itemize}[leftmargin=*,itemsep=1pt,topsep=2pt,parsep=0pt]
\item \textit{Feature attribution} (32 tests) asks which features matter: identifying the most important feature, ranking features, detecting irrelevant ones, and determining the sign of effects.
\item \textit{Point simulation} (43 tests) asks the LLM to predict the model's output for a specific input, ranging from simple single-feature queries to 20-feature sparse problems.
\item \textit{Sensitivity analysis} (32 tests) asks how predictions change when inputs change: direction of change, unit sensitivity, and nonlinear threshold detection.
\item \textit{Counterfactual reasoning} (28 tests) poses inverse problems: given a target output, find the input value that produces it.
\item \textit{Structural understanding} (28 tests) asks meta-questions about the model itself: whether it is compact and what its decision regions look like.
\item \textit{Complex function simulation} (37 tests) tests simulation on nonlinear synthetic data including interactions, piecewise functions, exponential decay, and nested thresholds.
\end{itemize}

The tests are randomly split into a development set (43 tests) and a held-out set (157 tests), which can be used to verify that the model strings are genuinely simulatable rather than simply reciting the answers to the original tests.
The full list of the tests with descriptions and pass rates is in \cref{tab:original_tests} (development set) and \cref{tab:gen_tests} (held-out set).

\subsection{Fully autonomous agentic autoresearch loop}
\label{sec:agent_loop}

Given the two evaluation metrics above, we use a coding agent to search the space of model implementations, similar to recent work on ``autoresearch''~\citep{novikov2025alphaevolve,romera2023mathematical,singh2023explaining,wang2025thetaevolve,yuksekgonul2026learning}\footnote{See sample implementation at \url{https://github.com/karpathy/autoresearch}.}.
We instruct the agent to modify only a single file, \texttt{interpretable\_regressor.py}, containing a starter \texttt{InterpretableRegressor} class.
It then follows an iterative loop (\cref{fig:overview}a): (1) edit the model code with a new idea, (2) run both evaluations, (3) check whether metrics improved, (4) save the resulting metrics / model code, and (5) repeat.
Throughout this autoresearch loop, the agent persists its memory through a csv file that records each model's name, basic idea, and resulting metrics.
Results for baseline models are recorded in this csv file before starting the loop, providing a reference point for the agent's search.

\paragraph{Prompting choices in the autoresearch loop.}
The agentic loop is highly configurable to various choices the user has in mind.
We prompt the coding agent to explore new ideas, i.e. the prompt ends with \textit{Do not simply import a known interpretable model and change its hyperparameters — build your own from scratch using basic building blocks or substantially modify an existing one. The goal is to discover new models, not just find the best hyperparameters for known models...BE CREATIVE!}

To ensure the agent continues, we prompt it to continue, i.e. \textit{Once the experiment loop has begun, do NOT pause to ask the human if you should continue. Run until manually stopped. Always keep going.
}. If the agent does stop, it is automatically prompted with the command \textit{Continue.} until the user has determined that sufficient iterations have been completed (in our experiments typically 50-200 iterations).
We additionally prompt the agent to avoid simply memorizing and reciting the answers to the given interpretability tests, i.e. \textit{Do not simply write the answers to the interpretability tests without actually building a model}.
See the full prompts used in the agentic loop in the Github repo.

The token usage for running the autoresearch loop varies by model and thinking effort; the total generated tokens across all our final experiments is approximately 70 million tokens (see detailed breakdown in \cref{tab:token_usage}). While this is a substantial number of tokens, the cost to replicate these experiments is relatively modest (as of Spring 2026, replicating the experiments that use Claude Code can be done with 3 months of a \$100 Claude Max subscription).

\section{Results}
\label{sec:results}

\subsection{Experimental setup}
\label{sec:exp_setup}

\paragraph{Datasets.}
We evaluate predictive performance during the \method loop on 65 regression datasets, consisting of all the regression datasets from the OpenML TabArena suite~\citep{erickson2025tabarena} (7 datasets) and all the regression datasets from PMLB excluding the duplicated synthetic Friedman datasets~\citep{olson2017pmlb} (58 datasets).
For evaluation on held-out data, we use the 16 OpenML regression datasets from a recent study~\citep{grinsztajn2022tree} that do not overlap with the development datasets.
Each dataset is preprocessed with an 80\%-20\% train-test split, subsampled to at most 1{,}000 samples and 50 features, and the outcome variable is normalized to zero mean and unit standard deviation on the training set.
Categorical features are ordinal-encoded and missing values are imputed with medians.
See details for all development datasets in \cref{tab:datasets} and held-out datasets in \cref{tab:gen_datasets}.

\paragraph{Baselines.}
We compare evolved models against 16 baselines spanning 5 families:
\textit{Linear:} linear regression (OLS), Ridge regression, the Lasso~\citep{tibshirani1996regression};
\textit{Tree:} DecisionTree with 8 and 20 leaf nodes (DT mini and DT large), HSTree with 8 and 20 leaf nodes (HSTree mini and HSTree large)~\citep{agarwal2022Hierarchical};
\textit{Additive:} PyGAM~\citep{hastie2017generalized}, Explainable boosting machine (EBM)~\citep{caruana2015intelligible,lou2013accurate};
\textit{Rule-based:} FIGS (8 and 20 rules)~\citep{tan2022Fast}, RuleFit~\citep{friedman2008predictive};
\textit{Black-box:} RandomForest (RF)~\citep{breiman2001random}, Gradient-boosted decision trees (GBM), multilayer perceptron (MLP), TabPFN-v2.5~\citep{hollmann2025accurate}.
These models are run with their default hyperparameters in their respective packages (\texttt{scikit-learn}~\citep{pedregosa2011scikit}, \texttt{imodels}~\citep{singh2021imodels}, \texttt{interpretML}~\citep{nori2019interpretml}), including default settings for selecting hyperparameters via cross-validation.
Each baseline provides a \texttt{\_\_str\_\_} method that renders its internal structure as text (e.g., tree diagrams, coefficient tables, or feature importance rankings).

\paragraph{Agent configurations.}
We run \method with two coding agents, Claude Code v2.1.118 (\texttt{Opus-4.6}/\texttt{Opus-4.7}) and Codex v0.118.0 (\texttt{GPT-5.3}), at 3 reasoning-effort levels each (medium, high, extra-high), for 6 runs total producing 42-103 working models per run.
We additionally run Claude Code \texttt{Opus-4.7} at medium reasoning effort 3 times, each with minor prompt variations, e.g. adding \textit{Be creative!} or \textit{Try out-of-the-box ideas!}, producing 44-63 working models per run.
All runs use \texttt{GPT-4o}~\citep{hurst2024gpt} as the LLM evaluator unless otherwise noted.

\paragraph{End-to-end ADS setup.}
% For end-to-end data analysis, we evaluate
To test whether the evolved models help in practice, we evaluate four AI agents on the BLADE benchmark~\citep{gu2024blade}, which consists of 13 datasets, each paired with a data-grounded question:
GitHub Copilot CLI v1.0.40 (powered by \texttt{Gemini-2.5-pro} or \texttt{Sonnet-4.5}), Claude Code v2.1.117 (\texttt{Sonnet-4.6}), and Codex v0.118.0 (\texttt{GPT-5.3}).
% 13 BLADE datasets, comparing two conditions.
For each question, BLADE provides a gold-standard analysis from expert data scientists and researchers.

We run BLADE with 4 conditions.
In the first condition (standard tools), the agent is prompted to use widespread interpretable models from \texttt{scikit-learn}, \texttt{imodels}, \texttt{statsmodels}, and \texttt{scipy.stats}.
In the second condition (\method), the agent is additionally given access to the evolved \method formatted as a Python package: we curate the library generated by top-performing models across our runs (the highlighted models in \cref{fig:overview}b) into a package with 10 evolved regressors spanning the interpretability-performance frontier, and point the agent to a \texttt{README.md} that documents the API and how to use the models.
The third and fourth conditions are designed to test whether any observed improvement in the \method condition is due to explicitly emphasizing a particular package rather than the substance of the models in that package.
They use the same prompt wording as in the \method condition, but change the line pointing to the \method package: in one control the prompt points to the \texttt{imodels} package~\citep{singh2021imodels} and in the other it points to the \texttt{interpretML} package~\citep{nori2019interpretml}.

Every ADS agent then provides analyses for each dataset under all four conditions.
Each resulting analysis is compared to the gold-standard human analysis using a rubric to assess (i) correctness, (ii) completeness, and (iii) clarity, each on a scale of 1 (worst) to 10 (best).
Each analysis is run 3 times and the resulting analysis is judged 3 times by \texttt{GPT-4o}, resulting in 9 evaluations per dataset per condition per agent; see prompts for the judge rubric and for running the agent on \href{https://anonymous.4open.science/r/agentic-imodels-apr25/}{GitHub}.

\subsection{\method push the interpretability-performance frontier}
\label{sec:main_results}

\begin{figure}[t]
    \centering
    \vspace{-20pt}
    \includegraphics[width=\textwidth]{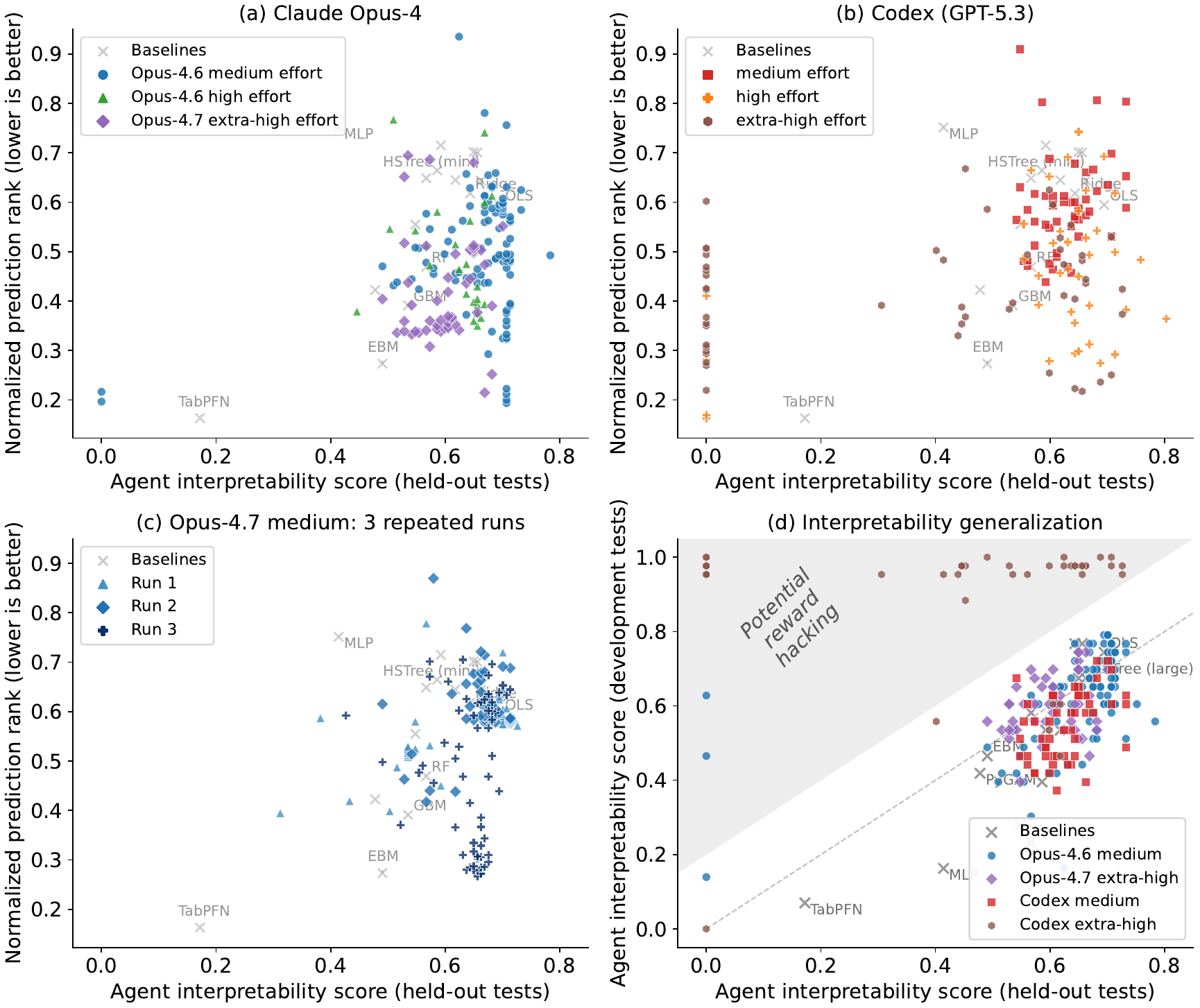}
    \caption{\method versus baselines (gray crosses) in terms of both predictive performance
    (the RMSE mean rank: each model's mean rank is computed across datasets, then normalized to $[0, 1]$ with lower being better) and agent interpretability scores (fraction of tests passed from the 157-test held-out generalization suite (\cref{tab:gen_tests})).
    Across different settings, \method achieve Pareto improvements in terms of predictive performance and interpretability:
    (a) Claude \texttt{Opus-4} models across three effort levels,
    (b) Codex (\texttt{GPT-5.3}) models across three effort levels,
    (c) Claude \texttt{Opus-4.7} at medium effort with 3 random repetitions.
    (d) Agent interpretability scores on the development set of tests versus the held-out set for four of the runs shown in (a)/(b) (using matching colors/markers).
    Some models exhibit held-out agent interpretability scores significantly below their development scores, suggesting potential reward hacking.
    Excluding the points in this shaded region, the remaining points show a strong positive correlation ($r=0.65$).}
    \label{fig:analysis}
\end{figure}

\cref{fig:analysis}a-c shows the interpretability-predictive performance frontier across the different models (breaking down the results shown in \cref{fig:overview}b).
Predictive performance is measured via the normalized prediction rank, i.e. we pool the models from all runs (16 baselines plus 467 evolved models from the 9 \method runs), rank them by RMSE on each of the 65 tabular datasets, average the per-dataset ranks for each model, and divide by the pool size so that ranks lie in $[0, 1]$ with lower being better.
Baselines (gray crosses) show a clear tradeoff: TabPFN (rank 0.16, agent interpretability score 0.17) predicts best but suffers from weak interpretability while the most interpretable baselines (OLS interpretability 0.69, HSTree-mini at 0.66) have normalized ranks of 0.60-0.82.

\paragraph{\method yield Pareto improvements.}
Evolved \method populate the previously empty low-rank, high-interpretability region across different coding agents and reasoning efforts.
For example, Claude Code (\cref{fig:analysis}a) produces \texttt{HingeEBM (5bag)} at rank 0.19 with agent interpretability score 0.71 (nearly matching TabPFN's rank at over 4 times its agent interpretability score).
Codex (\cref{fig:analysis}b) reaches comparable interpretability with models such as \texttt{TeacherStudentRuleSpline (v1)} (rank 0.36, agent interpretability score 0.80), and tends toward simpler sparse/linear modifications.
Even at a fixed agent and reasoning effort, runs display high variability (\cref{fig:analysis}c);
this variability is on par with the differences observed across different reasoning efforts and agents.

\paragraph{Reward hacking of development interpretability tests.}
\cref{fig:analysis}d compares each model's interpretability on the development interpretability tests versus the held-out tests.
Some evolved models (most prominently Codex extra-high) score high on development tests but substantially lower on held-out tests (shaded region).
Manually inspecting these models reveals that they often engage in reward hacking, e.g. reciting answers to the development interpretability tests directly in their model strings.
Nevertheless, most evolved models successfully generalize: after excluding the shaded region, the two agent interpretability scores are strongly correlated ($r = 0.65$, 228 points).

\paragraph{Validating the agent interpretability metric via  ablations.}
Beyond the held-out interpretability tests above, we ask whether the evolved models' gains are stable under two ablations of the evaluation pipeline.
First, we test perturbations to the evaluation of the agent interpretability tests.
We evaluate swapping the LLM evaluator from \texttt{GPT-4o} to \texttt{GPT-5.4} or Claude \texttt{Haiku-4.5}, also adding minor variations to the prompt used for evaluating tests.
We find that results are consistent across the evaluator LLM and the prompt variationsin all cases, including when using \texttt{GPT-5.4} for post-hoc evaluation (correlation of scores between evaluators is 0.83),
when Claude \texttt{Haiku-4.5} is used for post-hoc evaluation (correlation of 0.85),
 or when the evaluator is changed during the \method optimization loop (where the \method loop continues to achieve Pareto improvements); see \cref{app:gpt54} for extended results.

\paragraph{Validating predictive performance on held-out datasets.}
Second, we test whether the evolved models' predictive performance generalizes to held-out datasets.
We re-evaluate predictive performance for the 103 evolved models from the Claude \texttt{Opus-4.6} medium-effort main run (blue points in \cref{fig:analysis}a) on 16 held-out OpenML datasets in addition to the 157 new held-out interpretability tests, finding that the same conclusions hold and evolved models continue to dominate the Pareto frontier (see \cref{app:gen_scatter}).
The best-ranked evolved model, \texttt{SmartAdditive (v1)} yields 69\% agent interpretability (beating EBM's 49\% interpretability) while remaining competitive on rank, and the evolved Pareto frontier extends up to \texttt{HingeGAM (10bp)} at 78\% interpretability.

\subsection{End-to-end ADS performance improves using \method}
\label{sec:e2e_results}

\begin{figure}[t]
    \centering
    \vspace{-20pt}
    \makebox[\textwidth][c]{\includegraphics[width=1.3\textwidth]{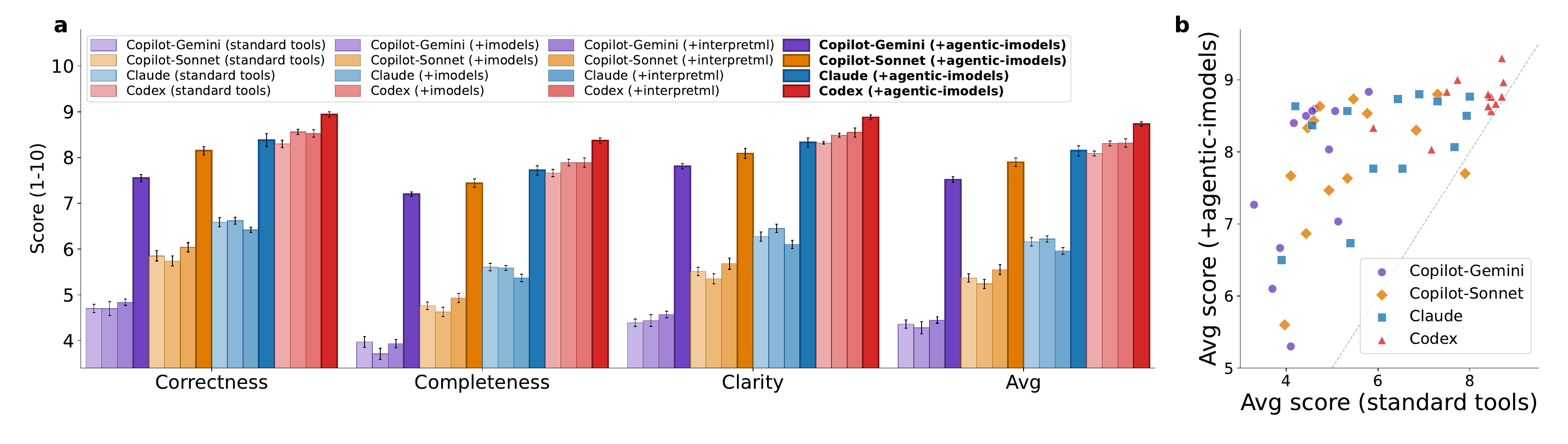}}
    \vspace{-20pt}
    \caption{
    Including \method improves performance on the BLADE benchmark across 4 ADS agents: GitHub Copilot CLI (\texttt{gemini-2.5-pro}), GitHub Copilot CLI (\texttt{sonnet-4.5}), Claude Code (\texttt{sonnet-4.6}), and Codex CLI (\texttt{GPT-5.3}).
    (a) Aggregate scores across the 13 BLADE datasets, with four prompt conditions per agent: standard tools (no explicit package emphasis), prompt emphasizing the \texttt{imodels} package, prompt emphasizing the \texttt{interpretML} package, and prompt emphasizing our \method package.
    Including \method yields substantial improvements over all the other conditions, across different agents and evaluation axes, particularly for the weaker ADS systems where the margin for improvement is larger.
    (b) Per-dataset average score (mean of correctness, completeness, clarity) with performance using standard tools versus \method; points above the diagonal indicate improvement.
    Error bars show standard error of the mean across agent seeds and judge seeds (9 evaluations per dataset per condition: 3 agent runs $\times$ 3 judge runs).
    }
    \label{fig:e2e}
\end{figure}

In the end-to-end ADS evaluation on the BLADE datasets, we find that providing every agent with the evolved \method improves its performance across all 3 rubric dimensions (\cref{fig:e2e}a).
Improvements are largest for the ADS systems with weaker base models than for the strongest one (Codex): Copilot CLI driven by \texttt{gemini-2.5-pro} improves its average score by $72.5\%$ with 13/13 datasets improved, Copilot CLI driven by \texttt{sonnet-4.5} improves by $47.0\%$ with 13/13 improved, Claude Code (\texttt{sonnet-4.6}) improves by $32.3\%$ with 13/13 improved, and Codex (\texttt{GPT-5.3}) improves by $7.9\%$ with 13/13 improved (\cref{fig:e2e}b).
% Agents with lower standard-tools baselines see the largest absolute gains, likely because they have a larger margin for improvement.
Explicitly emphasizing the \texttt{imodels} or \texttt{interpretML} packages in the prompt does not yield significant improvements for the weaker ADS systems (Copilot-Gemini shifts by $-1.8\%$/$+2.1\%$ and Copilot-Sonnet by $-2.4\%$/$+3.4\%$ for \texttt{imodels}/\texttt{interpretML}, all within the SE bars of the no-emphasis baseline) and yields a relatively small improvement for Codex ($2.7\%$ for \texttt{imodels} and $2.8\%$ for \texttt{interpretML} compared to $7.9\%$ for \method).

The gains in performance are attributable to different properties of the provided evolved models.
For example, one qualitative gain comes from the per-feature linear coefficients that \texttt{HingeEBM} prints in its \texttt{\_\_str\_\_}, which give the agent a compact magnitude-and-sign summary for every input.
On the \textit{mortgage} dataset~\citep{munnell1996mortgage}, this surfaces gender's small but non-zero effect (average score $4.6 \to 8.6$ Copilot-Gemini, $5.9 \to 7.8$ Claude, $7.2 \to 8.0$ Codex); on the \textit{Crofoot} dataset~\citep{crofoot2008interaction}, it lets the agent report a directional effect for specific variables rather than a binary significance judgement, raising completeness by $+2.7$ for both Codex and Claude.

\subsection{Qualitative analysis of selected \method}
\label{sec:qualitative_model_descriptions}

In this section we analyze patterns across 10 of the high-performing evolved models (the models highlighted in \cref{fig:overview}b).
See details and example \texttt{\_\_str\_\_} outputs for some of the models in \cref{app:model_descriptions}.

\paragraph{Pattern 1: bounded display complexity by construction.}
Across the 10 models, the agents often bound the size of \texttt{\_\_str\_\_} through hard architectural caps rather than soft regularization on a flexible model.
The caps take many forms: a fixed tree depth (e.g., \texttt{TinyDTDepth2}), a top-$k$ feature budget, a fixed number of quantile knots or breakpoints in a hinge basis, a forward-selected basis with rounded coefficients, or a single-line symbolic equation distilled from a stronger teacher.
% Across these examples, the agents commit upfront to a small structural budget for what will be displayed, then fit within that budget, rather than fitting freely and trimming after the fact.
This contrasts with strong baselines such as EBM, FIGS, and PyGAM, whose displays grow with the data and which sit at the worst end of the held-out interpretability axis despite predicting well (\cref{fig:overview}b).
This intrinsic interpretability principle echoes classical interpretable-ML works that treat complexity as a model constraint~\citep{rudin2018please,ustun2016supersparse,breiman1984classification}, and it keeps the printed string short enough for the LLM evaluator to simulate end-to-end.

\paragraph{Pattern 2: display optimization as a first-class design axis.}
Eight of the ten models pair a standard predictor with a non-trivial display strategy that is optimized separately from the architecture itself.
Examples include (i) per-feature gating that renders a learned shape function as a single coefficient when an $R^2$ test passes and a short piecewise table otherwise (e.g. \texttt{SmartAdditive}); (ii) collapsing a hinge or basis expansion into an effective linear slope for the printed equation while keeping the precise predictor underneath (e.g. \texttt{HingeEBM});
and (iii) compiling a distilled student into a single-row symbolic equation written specifically for \texttt{\_\_str\_\_}.
The choice of display also varies based on the underlying agent: Codex with \texttt{GPT-5.3} tends toward sparse symbolic displays, while Claude with \texttt{Opus-4.7} tends toward GAM-style displays.

\FloatBarrier
\section{Discussion}

Decades of interpretability research have optimized models for a human reader.
In the coming decades, with the increasing capabilities of AI agents, building tools for agents to reliably interpret data will be critical to ensuring that insights from data are correctly understood and acted upon.
\method is a first step toward taking this shift seriously, and may be the first of many works that adapt the structural properties that aid human interpretability (e.g., sparsity, additivity, monotonicity) into LLM-suitable properties.

\paragraph{Limitations.}
One limitation of the study here is that our end-to-end ADS evaluations rely on LLM-as-judge for scoring, which may introduce some bias or artifacts (although notably the underlying ground-truth analysis is conducted by expert humans, making the judging task easier).
% , which is sensitive to the choice of evaluator model (Appendix~\ref{app:gpt54}).
Similarly, our interpretability tests use LLMs for evaluation.
While this is effective for evaluating the agent-interpretability of the models being tested, it is unclear how well this measures the human-interpretability of representations, a question that may be explored in future user studies.
As LLM user simulators become more accurate~\citep{naous2025flipping,wu2026humanlm}, they could be used to automatically evaluate the interpretability of models to provide more reliable assessments that match human judgements.
Additionally, the interpretability tests we design do not cover all possible aspects of model interpretability, and particular applications may require specialized tests beyond our current suite.

Another limitation is the presence of reward hacking in the experiments, which highlights that our metric is not a perfect proxy. However, the strong correlation between held-out agent interpretability scores and downstream performance suggests that the metric captures useful signal despite being imperfect.
Future works may want to design the interpretability tests and the agentic loop to be more robust to increasingly sophisticated agents.
For example,
% interpretability tests could be programmatically re-generated on each iteration so the agent cannot memorize them.
% Alternatively,
tests could be exposed only through an API so the agent never sees the questions or expected answers in plain text.
Finally, the agent loop is expensive in LLM API calls, both for the coding agent and for the interpretability evaluator, although these costs may decrease as LLM inference becomes cheaper and more efficient over time.

\paragraph{Future work.}
A natural next step is to broaden the scope of the framework along two axes: across tasks (e.g. classification, time series, text data, larger datasets) and across more general tools (e.g. text explanation pipelines, causal interventions).
The same test suite could also be repurposed as an optimization target for improving existing model packages without re-architecting them~\citep{singh2021imodels,nori2019interpretml}.
Downstream, incorporating the evolved models into real-world agentic workflows (e.g., scientific discovery pipelines) would test whether the improvements transfer to more complex analyses.
On the human side, our framework could be modified to optimize human-AI collaborative workflows, rather than focusing on agents or humans in isolation~\citep{bansal2021does,feng2026human}.
As LLMs themselves improve at parsing structured models~\citep{lengerich2023llms}, \method will be increasingly powerful and both the evolved tools and the metric they optimize against will enable more sophisticated, potentially superhuman interpretation of complex data.
\FloatBarrier
{
    \small
    \bibliography{main.bib}
    \bibliographystyle{unsrt}
}
\FloatBarrier

\appendix
\counterwithin{figure}{section}
\counterwithin{table}{section}
\renewcommand{\thefigure}{A\arabic{figure}}
\renewcommand{\thetable}{A\arabic{table}}
\section{Experimental setup details}

\subsection{Interpretability test details}
\label{app:test_details}

\cref{tab:original_tests} lists all 43 development interpretability tests (the set used inside the \method optimization loop) with descriptions and pass rates pooled across all models (baselines + evolved candidates) from the 9 \method runs mentioned in \cref{sec:main_results}.
\cref{tab:gen_tests} lists the 157 held-out tests used for post-hoc evaluation, with pass rates pooled the same way.

\begin{table}[ht]
\centering
\caption{All 43 interpretability tests used inside the \method optimization loop, grouped by cognitive operation. Pass rates are pooled across the 9 \method runs mentioned in \cref{sec:main_results}; each run contributes its full set of evaluated models (baselines + evolved candidates).}
\label{tab:original_tests}
\scriptsize
\begin{tabular}{p{1.3cm}lp{5.0cm}r}
\toprule
Category & Test & Description & Pass \% \\
\midrule
\multirow{6}{1.3cm}{Feature\newline attribution} & most important feature & Identify the single most important feature & 97\% \\
 & feature ranking & Rank top 3 features by importance & 98\% \\
 & irrelevant features & Identify features with negligible effect & 87\% \\
 & sparse feature set & Identify active features in 10-feature data & 82\% \\
 & dominant feature sample & Identify dominant feature for a specific sample & 100\% \\
 & sign of effect & Determine sign and magnitude of a feature effect & 76\% \\
\midrule
\multirow{17}{1.3cm}{Point\newline simulation} & point prediction & Predict output for a specific input & 90\% \\
 & counterfactual prediction & Predict output for a different input & 89\% \\
 & all features active & Predict with all features non-zero & 66\% \\
 & two feature perturbation & Predict under simultaneous two-feature changes & 68\% \\
 & mixed sign goes negative & Handle opposite-signed features & 41\% \\
 & pairwise anti intuitive & Compare two complex samples & 13\% \\
 & predict above threshold & Predict a sample above a threshold & 72\% \\
 & predict below threshold & Predict a sample below a threshold & 85\% \\
 & simulate mixed sign & Trace 6 features with mixed-sign coefficients & 53\% \\
 & simulate all active & Simulate 5-feature linear with mixed signs & 27\% \\
 & simulatability & Simulate complex 4-feature prediction & 44\% \\
 & eight features & 8-feature linear mixed-sign & 46\% \\
 & fifteen features sparse & 15 features, only 3 active & 23\% \\
 & twenty features sparse & 20 features, only 4 active & 35\% \\
 & twelve features all active & 12 features all contributing & 24\% \\
 & quadratic & $y = 3x_0^2 - 2x_1^2 + x_2$ & 59\% \\
 & friedman1 & Classic Friedman-1 nonlinear benchmark & 63\% \\
\midrule
\multirow{6}{1.3cm}{Sensitivity\newline analysis} & direction of change & Direction of output change when a feature increases & 82\% \\
 & quantitative sensitivity & Quantify sensitivity over larger ranges & 88\% \\
 & unit sensitivity & Exact unit change prediction (tight tolerance) & 79\% \\
 & nonlinear direction & Predict output under nonlinearity & 66\% \\
 & threshold identification & Identify threshold values separating predictions & 75\% \\
 & nonlinear threshold & Detect hockey-stick/ReLU-like behavior & 52\% \\
\midrule
\multirow{2}{1.3cm}{Counterfactual\newline reasoning} & counterfactual target & Solve inverse problem: find input for target output & 25\% \\
 & quadratic counterfactual & Inverse problem on nonlinear function & 52\% \\
\midrule
\multirow{2}{1.3cm}{Structural\newline understanding} & compactness & Can the model be computed in $\leq$10 operations? & 81\% \\
 & decision region & Identify decision boundaries & 87\% \\
\midrule
\multirow{10}{1.3cm}{Complex fn.\newline simulation} & simulate double threshold & Two step-thresholds on one feature & 66\% \\
 & simulate additive nonlinear & Simulate $y = 3\max(0,x_0) + 2\sin(x_1) + x_2$ & 60\% \\
 & simulate interaction & Simulate $y = 3x_0 + 2x_1 + 1.5x_0 x_1$ & 50\% \\
 & triple interaction & Multi-way interactions & 79\% \\
 & cascading threshold & If-then cascading structure & 74\% \\
 & exponential decay & $y = 5\exp(-x_0) + 2x_1$ & 48\% \\
 & piecewise three segment & 3-segment piecewise linear & 83\% \\
 & sinusoidal & Trigonometric nonlinearity & 46\% \\
 & abs value & $y = 3|x_0| - 2|x_1| + x_2$ & 52\% \\
 & nested threshold & Nested if-then logic & 59\% \\
\bottomrule
\end{tabular}
\end{table}

\begin{table}[ht]
\centering
\caption{All 157 new interpretability tests used in the generalization experiment, aggregate.
 $N$ is the number of distinct tests in each sub-category; pass rates are pooled across the 9 \method runs mentioned in \cref{sec:main_results}.}
\label{tab:gen_tests}
\scriptsize
\begin{tabular}{p{1.3cm}lp{4.8cm}rr}
\toprule
Category & Sub-category & Description & $N$ & Pass \%\\
\midrule
\multirow{9}{1.3cm}{Feature\newline attribution\newline (26 tests)} & top feature & Most influential feature; answer varies across tests & 3 & 88\% \\
 & zero-effect feature & Identify the single inert feature & 3 & 7\% \\
 & top-2 ranking & List the two most influential features in order & 3 & 48\% \\
 & irrelevant set & Name every feature whose effect is negligible & 3 & 44\% \\
 & sign + unit effect & Signed change from a unit increase in a chosen feature & 3 & 75\% \\
 & dominant at sample & Dominant contributing feature for a given sample & 3 & 94\% \\
 & second feature & Identify the second-most-important feature & 3 & 77\% \\
 & count active & Count of features with non-negligible effect & 3 & 48\% \\
 & pair importance & Between two named features, which matters more & 2 & 89\% \\
\midrule
\multirow{9}{1.3cm}{Point\newline simulation\newline (26 tests)} & small linear & Predict output on 2--4 feature linear data & 3 & 29\% \\
 & all-features active & Predict with every feature non-zero (7--9 feats) & 3 & 90\% \\
 & sparse many-feature & Predict on 10--18 feats with only 3--4 active & 3 & 58\% \\
 & alternating sign & Predict when coefficients alternate in sign & 3 & 51\% \\
 & tail query & Predict at feature values with $|x_i| \geq 1.6$ & 3 & 60\% \\
 & irrational inputs & Predict at irregular, non-round query points & 3 & 47\% \\
 & partial input & Predict when only a subset of features is specified & 3 & 59\% \\
 & non-zero intercept & Predict under a shifted intercept & 3 & 26\% \\
 & compare two points & Compare predictions at two input samples & 2 & 50\% \\
\midrule
\multirow{9}{1.3cm}{Sensitivity\newline analysis\newline (26 tests)} & direction of change & Signed delta for a one-unit change at varied bases & 3 & 88\% \\
 & wide-range & Quantify sensitivity over a large interval & 3 & 62\% \\
 & tight unit change & Unit change with tight numerical tolerance & 3 & 76\% \\
 & crossing point & Find $x_k$ at which prediction crosses a target & 3 & 81\% \\
 & two-feature change & Delta when two features change simultaneously & 3 & 27\% \\
 & decrease phrasing & Signed delta phrased as a decrease & 3 & 33\% \\
 & multi-unit step & Delta under a multi-unit change & 3 & 19\% \\
 & non-zero base & Unit change from a non-zero baseline & 3 & 70\% \\
 & small step & Sub-unit step sensitivity & 2 & 95\% \\
\midrule
\multirow{9}{1.3cm}{Counterfactual\newline reasoning\newline (26 tests)} & inverse on linear & Value of $x_k$ that attains a target prediction & 3 & 64\% \\
 & inverse, varied feat & Counterfactual targeting a rotated feature & 3 & 60\% \\
 & inverse, sparse data & Counterfactual with many inert features & 3 & 26\% \\
 & inverse, nonlinear & Counterfactual on ReLU/quad/piecewise/abs data & 3 & 63\% \\
 & with intercept & Counterfactual under a non-zero intercept & 3 & 24\% \\
 & reverse direction & Counterfactual requiring a downward change & 3 & 35\% \\
 & large change & Counterfactual requiring a sizeable feature move & 3 & 55\% \\
 & mid-index feature & Counterfactual on a feature with middle index & 3 & 54\% \\
 & negative target & Counterfactual to a negative target prediction & 2 & 42\% \\
\midrule
\multirow{9}{1.3cm}{Structural\newline understanding\newline (26 tests)} & decision region & Threshold of $x_k$ for a specified prediction level & 3 & 71\% \\
 & compactness & Can the model be evaluated in $\leq N$ operations? & 3 & 67\% \\
 & argmax & Value of $x_k$ that maximises prediction & 3 & 33\% \\
 & argmin & Value of $x_k$ that minimises prediction & 3 & 34\% \\
 & monotonic direction & Monotonic direction of a chosen feature & 3 & 56\% \\
 & output range & Approximate range of the model's predictions & 3 & 45\% \\
 & decision mid-output & Threshold of $x_k$ for a mid-range prediction & 3 & 76\% \\
 & argmax nonlinear & Argmax on a nonlinear data-generating process & 3 & 46\% \\
 & plateau region & Region where predictions are approximately constant & 2 & 86\% \\
\midrule
\multirow{9}{1.3cm}{Complex fn.\newline simulation\newline (27 tests)} & quadratic & Simulate $y = c_0 + c_1 x_k + c_2 x_k^2$ & 3 & 92\% \\
 & pairwise interaction & Simulate linear $+ x_i \cdot x_j$ interaction & 3 & 74\% \\
 & triple interaction & Simulate multi-way feature products & 3 & 93\% \\
 & Friedman-1 & Simulate the Friedman-1 nonlinear benchmark & 3 & 22\% \\
 & cascading threshold & If-then cascading data-generating process & 3 & 84\% \\
 & exponential decay & Simulate $y = c \exp(-x_k) + \mathrm{linear}$ & 3 & 25\% \\
 & 3-segment piecewise & Simulate 3-segment piecewise linear function & 3 & 91\% \\
 & sinusoidal & Simulate trigonometric nonlinearity & 3 & 89\% \\
 & absolute value & Simulate $y = c_1 |x_i| + c_2 |x_j| + \mathrm{linear}$ & 3 & 92\% \\
\bottomrule
\end{tabular}
\end{table}

\subsection{Dataset details}
\label{app:datasets}

\cref{tab:datasets} lists all 65 datasets used in the main experiment, including 7 from OpenML and 58 from PMLB~\citep{olson2017pmlb}.
Each dataset is subsampled to at most 1{,}000 samples and 50 features during evaluation.
\cref{tab:gen_datasets} lists the 16 held-out OpenML datasets used in the generalization experiment (\cref{app:gen_scatter}).

% Total datasets: 65
\begin{table}[ht]
\centering
\caption{The 65 datasets used in the main experiment (7 OpenML, 58 PMLB). $n$ = samples, $p$ = features (before subsampling).}
\label{tab:datasets}
\scriptsize
\begin{tabular}{lrr@{\hskip 12pt}|lrr}
\toprule
Dataset & $n$ & $p$ & Dataset & $n$ & $p$ \\
\midrule
abalone & 4176 & 8 & 4544\_GeographicalOr.. & 1059 & 117 \\
california & 20634 & 8 & 485\_analcatdata\_ve.. & 48 & 4 \\
cpu\_act & 8192 & 21 & 503\_wind & 6574 & 14 \\
elevators & 16599 & 16 & 505\_tecator & 240 & 124 \\
house\_16H & 22784 & 16 & 519\_vinnie & 380 & 2 \\
kin8nm & 8192 & 8 & 522\_pm10 & 500 & 7 \\
pol & 15000 & 26 & 523\_analcatdata\_ne.. & 100 & 2 \\
1027\_ESL & 488 & 4 & 527\_analcatdata\_el.. & 67 & 14 \\
1028\_SWD & 1000 & 10 & 529\_pollen & 3848 & 4 \\
1029\_LEV & 1000 & 4 & 537\_houses & 20640 & 8 \\
1030\_ERA & 1000 & 4 & 542\_pollution & 60 & 15 \\
1089\_USCrime & 47 & 13 & 547\_no2 & 500 & 7 \\
1096\_FacultySalaries & 50 & 4 & 556\_analcatdata\_ap.. & 475 & 3 \\
1191\_BNG\_pbc & 1000000 & 18 & 557\_analcatdata\_ap.. & 475 & 3 \\
1193\_BNG\_lowbwt & 31104 & 9 & 560\_bodyfat & 252 & 14 \\
1196\_BNG\_pharynx & 1000000 & 10 & 561\_cpu & 209 & 7 \\
1199\_BNG\_echoMonths & 17496 & 9 & 562\_cpu\_small & 8192 & 12 \\
1201\_BNG\_breastTumor & 116640 & 9 & 564\_fried & 40768 & 10 \\
1203\_BNG\_pwLinear & 177147 & 10 & 573\_cpu\_act & 8192 & 21 \\
1595\_poker & 1025010 & 10 & 574\_house\_16H & 22784 & 16 \\
192\_vineyard & 52 & 2 & 659\_sleuth\_ex1714 & 47 & 7 \\
197\_cpu\_act & 8192 & 21 & 663\_rabe\_266 & 120 & 2 \\
201\_pol & 15000 & 48 & 665\_sleuth\_case2002 & 147 & 6 \\
210\_cloud & 108 & 5 & 666\_rmftsa\_ladata & 508 & 10 \\
215\_2dplanes & 40768 & 10 & 678\_visualizing\_en.. & 111 & 3 \\
218\_house\_8L & 22784 & 8 & 687\_sleuth\_ex1605 & 62 & 5 \\
225\_puma8NH & 8192 & 8 & 690\_visualizing\_ga.. & 323 & 4 \\
227\_cpu\_small & 8192 & 12 & 695\_chatfield\_4 & 235 & 12 \\
228\_elusage & 55 & 2 & 706\_sleuth\_case1202 & 93 & 6 \\
229\_pwLinear & 200 & 10 & 712\_chscase\_geyser1 & 222 & 2 \\
230\_machine\_cpu & 209 & 6 & banana & 5300 & 2 \\
294\_satellite\_image & 6435 & 36 & titanic & 2099 & 8 \\
344\_mv & 40768 & 10 & & & \\
\bottomrule
\end{tabular}
\end{table}

\begin{table}[ht]
\centering
\caption{The 16 held-out OpenML datasets (suite 335) used in the generalization experiment. $n$ = samples, $p$ = features (before subsampling).}
\label{tab:gen_datasets}
\small
\begin{tabular}{rlrr@{\hskip 12pt}|rlrr}
\toprule
ID & Dataset & $n$ & $p$ & ID & Dataset & $n$ & $p$ \\
\midrule
44055 & analcatdata\_sup.. & 4052 & 7 & 44068 & particulate-matt.. & 394299 & 6 \\
44056 & visualizing\_soil & 8641 & 4 & 44069 & SGEMM\_GPU\_kern.. & 241600 & 9 \\
44059 & diamonds & 53940 & 9 & 45041 & topo\_2\_1 & 8885 & 255 \\
44061 & Mercedes\_Benz\_.. & 4209 & 359 & 45043 & seattlecrime6 & 52031 & 4 \\
44062 & Brazilian\_houses & 10692 & 11 & 45045 & delays\_zurich\_.. & 5465575 & 11 \\
44063 & Bike\_Sharing\_D.. & 17379 & 11 & 45046 & Allstate\_Claims.. & 188318 & 124 \\
44065 & nyc-taxi-green-d.. & 581835 & 16 & 45047 & Airlines\_DepDel.. & 1000000 & 5 \\
44066 & house\_sales & 21613 & 17 & 45048 & medical\_charges & 163065 & 3 \\
\bottomrule
\end{tabular}
\end{table}

\FloatBarrier
\section{Extended results and ablations}
\counterwithin{figure}{section}
\counterwithin{table}{section}
\renewcommand{\thefigure}{B\arabic{figure}}
\renewcommand{\thetable}{B\arabic{table}}

\FloatBarrier

\subsection{Ablation: alternative LLM evaluators for the interpretability tests}
\label{app:gpt54}

\begin{figure}[ht]
    \centering
    \begin{minipage}[b]{0.583\textwidth}
        \includegraphics[width=\textwidth]{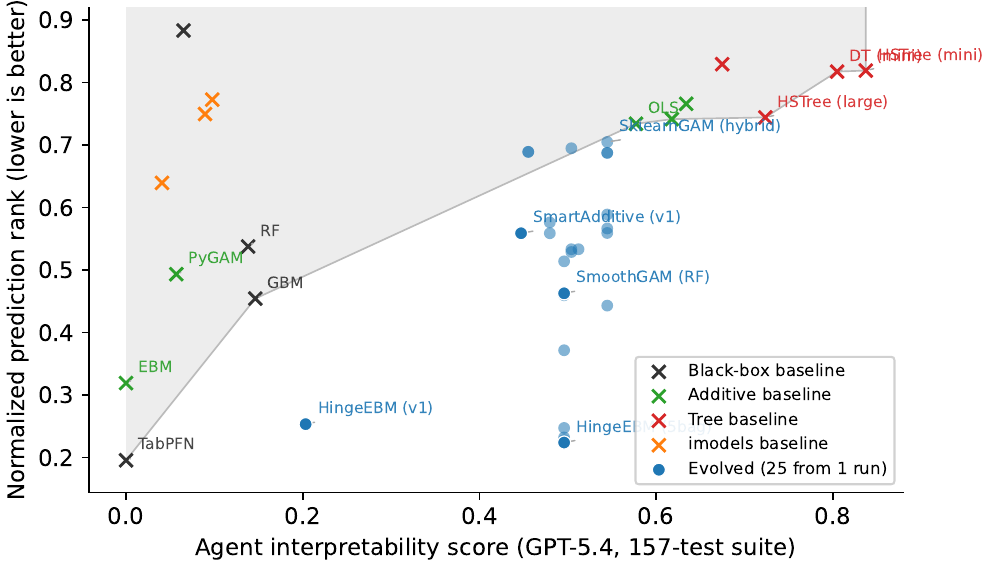}%
        \put(-215,130){\textbf{a}}%
    \end{minipage}%
    \hfill
    \begin{minipage}[b]{0.417\textwidth}
        \includegraphics[width=\textwidth]{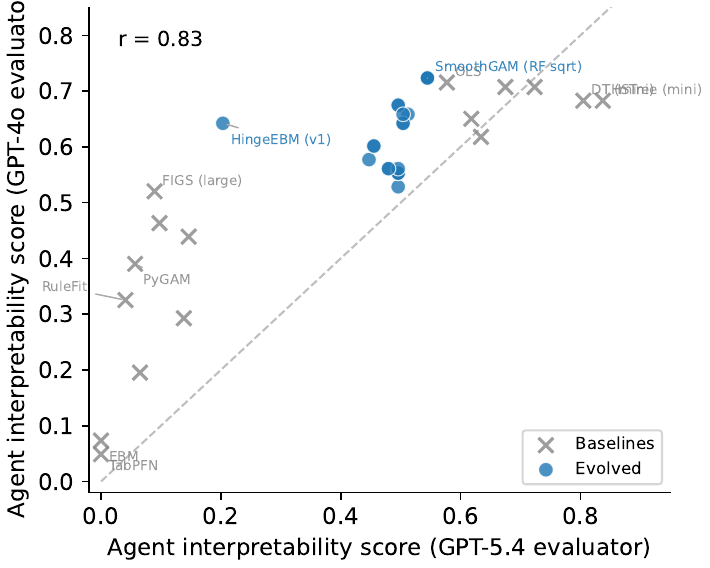}%
        \put(-160,130){\textbf{b}}%
    \end{minipage}\\[0.5ex]
    \begin{minipage}[b]{0.583\textwidth}
        \includegraphics[width=\textwidth]{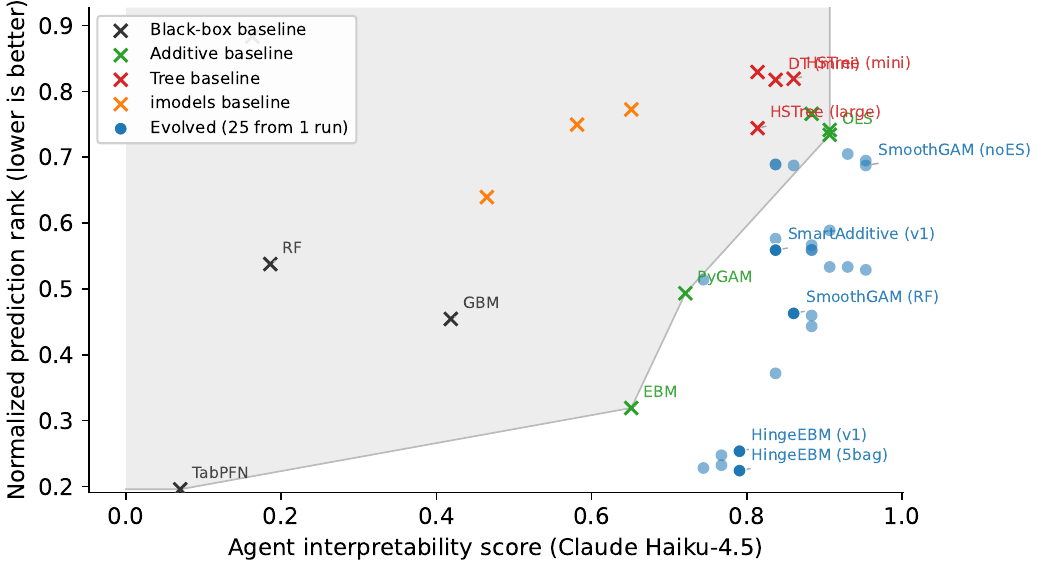}%
        \put(-215,130){\textbf{c}}%
    \end{minipage}%
    \hfill
    \begin{minipage}[b]{0.417\textwidth}
        \includegraphics[width=\textwidth]{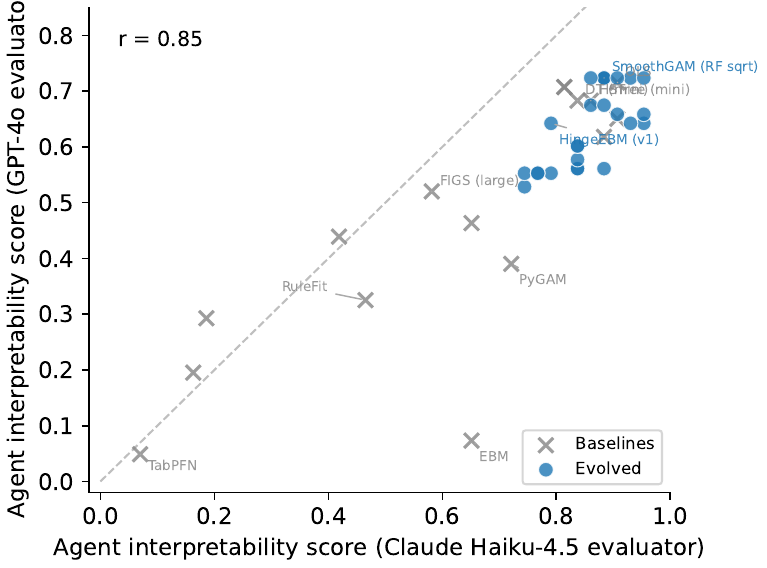}%
        \put(-160,130){\textbf{d}}%
    \end{minipage}
    \caption{Post-hoc evaluator-sensitivity analysis on the held-out interpretability tests (41 baseline + evolved models from the Claude \texttt{Opus-4.6} medium-effort run). \textbf{Top row}: \texttt{GPT-5.4} evaluator (same prompt template as the in-loop \texttt{GPT-4o} evaluator). \textbf{Bottom row}: Claude \texttt{Haiku-4.5} evaluator with a slightly perturbed prompt.
    (a, c) Normalized prediction rank on the 65 original training datasets (y-axis matching \cref{fig:analysis}a--c) versus agent interpretability score under each evaluator; baselines are colored crosses by category and evolved models are blue. The evolved Pareto frontier is preserved under both evaluators.
    (b, d) Per-model agent interpretability score under each held-out evaluator (x-axis) versus the original \texttt{GPT-4o} score (y-axis); baselines are gray, evolved models are blue.
    Scores correlate strongly with \texttt{GPT-4o} under both evaluators ($r = 0.83$ for \texttt{GPT-5.4}, $r = 0.85$ for Haiku), but with opposite biases: \texttt{GPT-5.4} is the stricter evaluator (most points above the diagonal), whereas Haiku is more lenient (most points below the diagonal).}
    \label{fig:gpt54_posthoc}
\end{figure}

\paragraph{\texttt{GPT-5.4} evaluator used post-hoc after models have been optimized.}
\label{app:gpt54_posthoc}
We reevaluate the 41 baseline + evolved models from the Claude \texttt{Opus-4.6} medium-effort main run under \texttt{GPT-5.4} instead of \texttt{GPT-4o} on the 43 development-set interpretability tests.
Only the LLM evaluator changes, so prediction ranks are unaffected.
\cref{fig:gpt54_posthoc} shows two views of the same 41-model pool.
\cref{fig:gpt54_posthoc}a shows where each model lands in the predictive performance versus interpretability plot under \texttt{GPT-5.4} (predictive performance is the same as is used in \cref{fig:analysis}a-c).
Despite the score compression for baselines (EBM and TabPFN drop to $\approx 0$; GBM from $0.53$ to $0.15$), the evolved Pareto frontier is preserved: \texttt{HingeEBM (5bag)} retains a rank of $13$ at $50\%$ interpretability, well outside the baseline Pareto region.

\cref{fig:gpt54_posthoc}b plots each model's \texttt{GPT-5.4} score (x-axis) against its \texttt{GPT-4o} score (y-axis).
The two scores are strongly correlated ($r = 0.83$), and within-evaluator consistency across three random seeds is high (standard deviation $<$ 0.03 for both evaluators), so the between-evaluator divergence reflects genuine differences in evaluation criteria rather than noise.
Most points fall above the diagonal, indicating \texttt{GPT-5.4} is the stricter evaluator overall, with the largest drops on baselines that have complex string representations (EBM, FIGS large, PyGAM, RuleFit) and the largest evolved-model drop on \texttt{HingeEBM (v1)}; simple decision trees (DT mini, HSTree mini) lie below the diagonal.
The drops are often attributable to \texttt{GPT-5.4} correctly abstaining from answering when the information is not present whereas \texttt{GPT-4o} would sometimes attempt to answer regardless, occasionally correctly answering some tests by chance.

\paragraph{Claude \texttt{Haiku-4.5} evaluator used post-hoc after models have been optimized.}
To check that the post-hoc results are not specific to a single evaluator family or to the exact prompt template used for evaluation,
we additionally re-evaluate the same 41-model pool with Claude \texttt{Haiku-4.5} and a slightly \emph{perturbed prompt} (re-worded preamble, explicit \texttt{MODEL:}/\texttt{QUESTION:} separators, and an instruction to use only the printed model).
\cref{fig:gpt54_posthoc}c-d show the result:
again, \method expands the Pareto frontier, and Haiku scores correlate with \texttt{GPT-4o} at $r = 0.85$.
This is comparable to the \texttt{GPT-5.4}-vs-\texttt{GPT-4o} correlation, although now most models lie below the diagonal.

\begin{figure}[ht]
    \centering
    \includegraphics[width=0.7\textwidth]{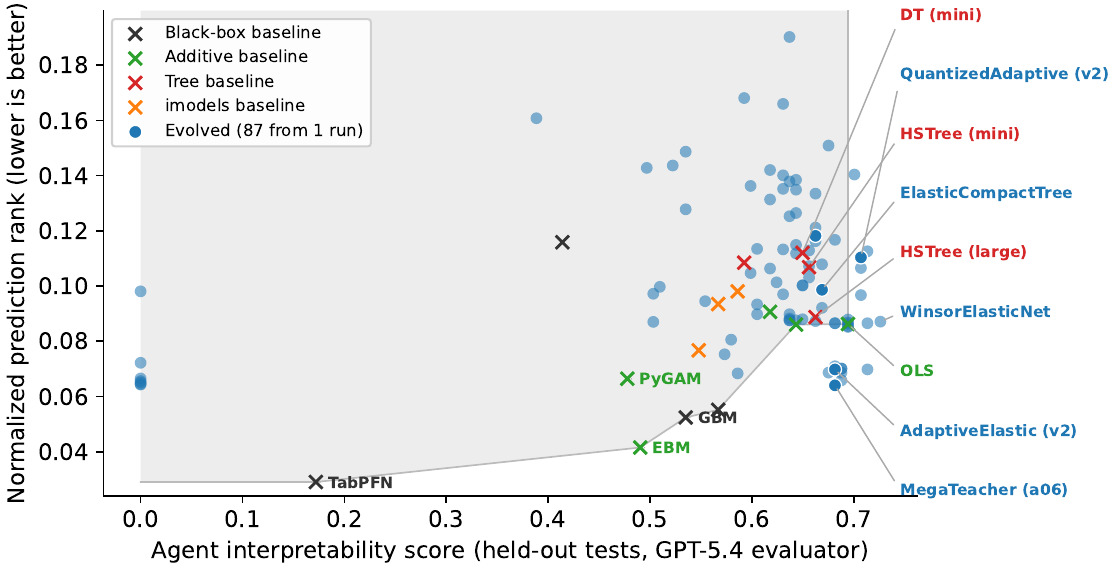}
    \caption{In-loop \texttt{GPT-5.4} optimization run: normalized prediction rank on the 65 training datasets versus agent interpretability score on the 157 held-out tests graded by \texttt{GPT-5.4}. The evolved models (blue) continue to improve over the Pareto frontier defined by the baselines (colored crosses).}
    \label{fig:gpt54_inloop}
\end{figure}

\FloatBarrier
\paragraph{\texttt{GPT-5.4} evaluator used during the optimization loop.}
\label{app:gpt54_optimization}
As a stronger test of evaluator stability, we also run the full \method optimization loop with \texttt{GPT-5.4} (rather than \texttt{GPT-4o}) as the in-loop interpretability evaluator. We run with Claude Code \texttt{Opus-4.6}, medium reasoning effort, which yields 87 evolved models.
\cref{fig:gpt54_inloop} shows the models (and the 16 baselines) on the same rank versus interpretability axes as \cref{fig:gpt54_posthoc}a, with interpretability measured on the held-out tests (also graded by \texttt{GPT-5.4}).

We find that the evolved models extend the Pareto frontier under the stricter evaluator: the best evolved models reach agent interpretability scores in the $0.71$ to $0.73$ range (e.g., \texttt{WinsorElasticNet} at normalized rank $0.09$ and interp $0.73$, \texttt{StepByStepAdaptive} at rank $0.07$ and interp $0.71$), above the best baseline (\texttt{OLS} at rank $0.09$ / interp $0.69$), while \texttt{MegaTeacher\_a06} (rank $0.06$ / interp $0.68$) matches or improves on every additive baseline in both dimensions.
The \method identified with the \texttt{GPT-5.4} evaluator differ from the \texttt{GPT-4o}-driven main runs: rather than hinge-based GAMs and additive boosted stumps, the \texttt{GPT-5.4} loop favors a mix of regularized linear models (\texttt{WinsorElasticNet}, \texttt{SparseBayesRidge}, \texttt{BayesianRidgeLinear}), per-dataset adaptive selectors (\texttt{AdaptiveElastic}, \texttt{QuantizedAdaptive}), and compact distilled trees (\texttt{MegaTeacher}, \texttt{TinyRecalTree}, \texttt{ElasticCompactTree}).
% , consistent with GPT-5.4's preference for simple functional forms (\cref{fig:gpt54_posthoc}b).
% The framework therefore adapts to evaluator choice both by preserving the existing frontier (\cref{app:gpt54_posthoc}) and by discovering new evaluator-appropriate architectures when re-optimized end to end.

\FloatBarrier
\subsection{Ablation: using held-out datasets to evaluate predictive performance}
\label{app:gen_scatter}

\begin{figure}[ht]
    \centering
    \begin{minipage}[b]{0.583\textwidth}
        \includegraphics[width=\textwidth]{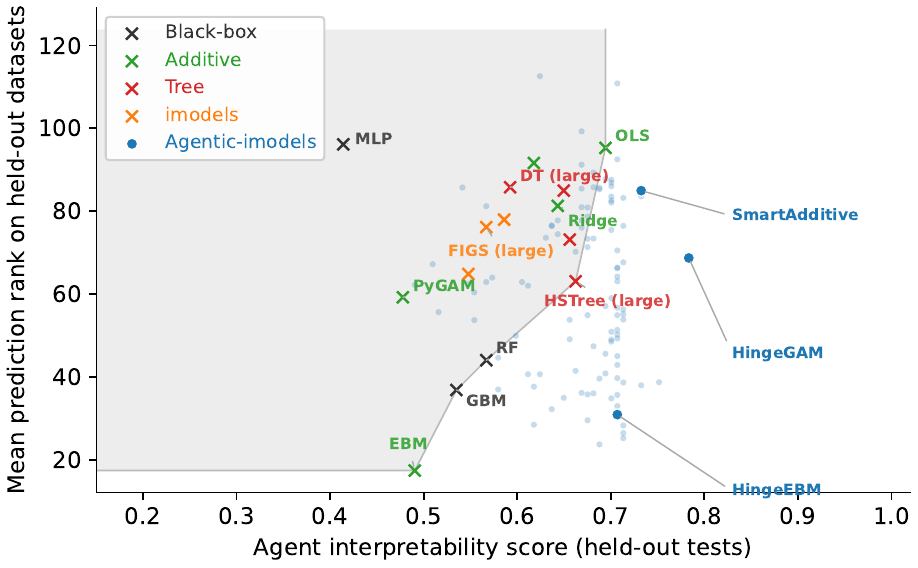}%
        \put(-200,150){\textbf{a}}%
    \end{minipage}%
    \hfill
    \begin{minipage}[b]{0.417\textwidth}
        \includegraphics[width=\textwidth]{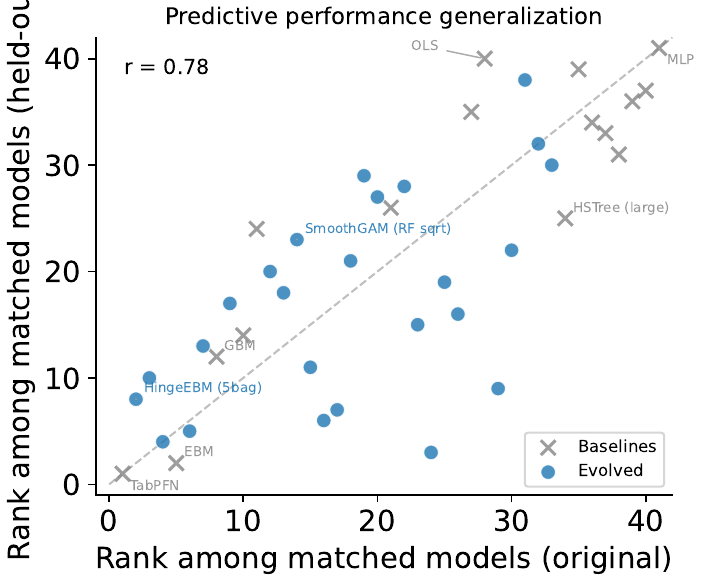}%
        \put(-145,150){\textbf{b}}%
    \end{minipage}
    \caption{(a) Interpretability versus prediction rank on 16 held-out datasets and 157 new interpretability tests for the Claude \texttt{Opus-4.6} medium-effort main run (103 evolved models, blue) together with the 16 baselines. Ranks are computed globally across this 119-model pool. Evolved families such as hinge-based (HingeEBM, HingeGAM) and smooth additive (SmartAdditive) models continue to dominate the Pareto frontier; selected class labels correspond to the highlighted models in \cref{fig:overview}b. (b) Prediction ranks on the original versus held-out evaluation for the 41 matched models, re-ranked independently within each evaluation. Ranks correlate at $r = 0.78$; variance is higher than for the interpretability comparison (\cref{fig:analysis}d) because the held-out datasets have different characteristics.}
    \label{fig:generalization}
\end{figure}

\FloatBarrier

\subsection{Extended results on the BLADE end-to-end ADS analysis}
\label{app:blade}

\cref{tab:blade_perdataset} shows the per-dataset scores for the BLADE end-to-end evaluation.

\newcommand{\se}[1]{{\tiny$\pm$#1}}

\begin{table}[h]
\centering
\small
\caption{Per-dataset BLADE average scores breaking down the results in \cref{fig:e2e}.
Scores are averaged over the three rubric dimensions and error bars show the standard error over 9 evaluations per agent per condition.
Std = standard interpretable tools, \method = with the agentic-imodels package. Bold indicates the higher score per agent when error bars do not overlap.}
\label{tab:blade_perdataset}
\setlength{\tabcolsep}{3pt}
\resizebox{\textwidth}{!}{%
\begin{tabular}{l cc cc cc cc}
\toprule
& \multicolumn{2}{c}{Copilot (Gemini)} & \multicolumn{2}{c}{Copilot (Sonnet)} & \multicolumn{2}{c}{Claude Code} & \multicolumn{2}{c}{Codex CLI} \\
\cmidrule(lr){2-3} \cmidrule(lr){4-5} \cmidrule(lr){6-7} \cmidrule(lr){8-9}
Dataset & Std & \method & Std & \method & Std & \method & Std & \method \\
\midrule
affairs         & 2.9\se{0.2} & \textbf{5.9}\se{0.8} & 4.4\se{0.1} & \textbf{6.9}\se{0.5} & 5.4\se{0.5} & 6.7\se{0.9} & 7.7\se{0.5} & \textbf{9.0}\se{0.2} \\
amtl            & 4.4\se{0.2} & \textbf{8.5}\se{0.2} & 6.9\se{0.6} & \textbf{8.3}\se{0.2} & 6.4\se{0.5} & \textbf{8.7}\se{0.0} & 8.6\se{0.2} & 8.7\se{0.1} \\
boxes           & 3.9\se{0.3} & \textbf{6.7}\se{0.8} & 4.1\se{0.4} & \textbf{7.7}\se{0.4} & 4.2\se{0.4} & \textbf{8.6}\se{0.1} & 8.7\se{0.1} & \textbf{9.0}\se{0.1} \\
caschools       & 5.0\se{0.2} & \textbf{8.0}\se{0.2} & 7.9\se{0.2} & 7.7\se{0.4} & 7.7\se{0.4} & 8.1\se{0.2} & 7.5\se{0.3} & \textbf{8.9}\se{0.1} \\
crofoot         & 3.3\se{0.3} & \textbf{7.3}\se{0.7} & 4.0\se{0.2} & \textbf{5.6}\se{0.4} & 3.9\se{0.2} & \textbf{6.5}\se{0.6} & 5.9\se{0.4} & \textbf{8.3}\se{0.2} \\
fertility       & 5.1\se{0.8} & \textbf{7.0}\se{0.7} & 4.7\se{0.3} & \textbf{8.6}\se{0.2} & 7.9\se{0.2} & \textbf{8.5}\se{0.1} & 8.4\se{0.1} & \textbf{8.8}\se{0.2} \\
fish            & 4.2\se{0.3} & \textbf{8.4}\se{0.1} & 5.4\se{0.2} & \textbf{8.7}\se{0.1} & 5.3\se{0.1} & \textbf{8.6}\se{0.1} & 8.4\se{0.2} & \textbf{8.6}\se{0.1} \\
hurricane       & 5.1\se{0.5} & \textbf{8.6}\se{0.1} & 5.8\se{0.8} & \textbf{8.5}\se{0.2} & 7.3\se{0.3} & \textbf{8.7}\se{0.1} & 8.7\se{0.1} & 8.8\se{0.1} \\
mortgage        & 4.6\se{0.3} & \textbf{8.6}\se{0.1} & 5.3\se{0.7} & \textbf{7.6}\se{0.3} & 5.9\se{0.3} & \textbf{7.8}\se{0.4} & 7.2\se{0.5} & 8.0\se{0.4} \\
panda\_nuts     & 4.6\se{0.2} & \textbf{8.6}\se{0.0} & 4.5\se{0.1} & \textbf{8.3}\se{0.2} & 6.5\se{0.6} & \textbf{7.8}\se{0.1} & 8.4\se{0.1} & \textbf{8.8}\se{0.1} \\
reading         & 3.7\se{0.2} & \textbf{6.1}\se{0.6} & 4.6\se{0.5} & \textbf{8.4}\se{0.2} & 4.6\se{0.3} & \textbf{8.4}\se{0.1} & 8.7\se{0.1} & \textbf{9.3}\se{0.1} \\
soccer          & 4.1\se{0.4} & \textbf{5.3}\se{0.7} & 5.0\se{0.1} & \textbf{7.5}\se{0.5} & 6.9\se{0.3} & \textbf{8.8}\se{0.1} & 8.4\se{0.1} & 8.6\se{0.1} \\
teachingratings & 5.8\se{0.5} & \textbf{8.8}\se{0.1} & 7.3\se{0.4} & \textbf{8.8}\se{0.1} & 8.0\se{0.2} & \textbf{8.8}\se{0.1} & 8.5\se{0.1} & \textbf{8.8}\se{0.1} \\
\midrule
\textbf{Average} & 4.36\se{0.09} & \textbf{7.52}\se{0.06} & 5.37\se{0.09} & \textbf{7.90}\se{0.09} & 6.16\se{0.09} & \textbf{8.15}\se{0.11} & 8.09\se{0.05} & \textbf{8.73}\se{0.05} \\
\bottomrule
\end{tabular}%
}
\end{table}

\FloatBarrier
\subsection{Token consumption}
\label{app:tokens}

\cref{tab:token_usage} summarizes approximate token consumption per \method run. We split the agent column into \emph{Fresh} tokens (input the model processes for the first time, plus cache-creation writes and outputs) and \emph{Cache reads} (the same conversational context being re-read from the prompt cache on each follow-up turn): cache reads dominate the raw byte count of an agentic loop but are billed at roughly $10\times$ less per token than fresh input on both Anthropic and OpenAI APIs, so the two columns are not directly comparable as ``compute consumed.'' Interpretability-test token counts are estimated by multiplying the number of \texttt{GPT-4o} calls (one row per test per candidate model) by an average of 1{,}050 tokens per call (prompt plus short response).

\begin{table}[ht]
\centering
\small
\caption{Approximate token consumption per \method run, broken down by what the model actually processed for the first time (\emph{Fresh}) versus what was re-read from the prompt cache on follow-up turns (\emph{Cache reads}). \emph{Fresh} sums fresh input, cache-creation, and output tokens (Anthropic) or non-cached input plus output (OpenAI); \emph{Cache reads} is the cache-replay traffic, which is billed at $\approx$10$\times$ less per token and so dominates the raw token totals without representing comparable compute. Interpretability-test tokens are estimated assuming $\approx$1{,}050 tokens per \texttt{GPT-4o} call (prompt plus response).
For the \texttt{GPT-5.4} evaluator run, we double the per-call estimate to $\approx$2{,}100 tokens to account for the larger reasoning traces produced by \texttt{GPT-5.4}.}
\label{tab:token_usage}
\noindent\makebox[\textwidth][c]{%
\begin{tabular}{l|rrr|rr|r}
\toprule
Run & Models & Fresh & Cache reads & Interp.\ calls & Interp.\ tokens & Total \\
 & tried & (M) & (M) & & (M) & (M) \\
\midrule
Claude \texttt{Opus-4.6} (medium) & 165 & 7.4 & 749.7 & 7,611 & 8.0 & 765.1 \\
Claude \texttt{Opus-4.7} (medium, run 1) & 46 & 0.8 & 34.8 & 2,666 & 2.8 & 38.4 \\
Claude \texttt{Opus-4.7} (medium, run 2) & 50 & 0.5 & 29.3 & 2,795 & 2.9 & 32.7 \\
Claude \texttt{Opus-4.7} (medium, run 3) & 64 & 1.2 & 54.2 & 3,440 & 3.6 & 59.0 \\
Claude \texttt{Opus-4.6} (high) & 87 & 1.2 & 185.4 & 4,214 & 4.4 & 191.0 \\
Claude \texttt{Opus-4.7} (xhigh) & 65 & 4.1 & 212.2 & 3,569 & 3.7 & 220.0 \\
Claude \texttt{Opus-4.6} (medium, \texttt{GPT-5.4} evaluator) & 89 & 7.6 & 803.7 & 4,515 & 9.4 & 820.7 \\
Codex \texttt{GPT-5.3} (default) & 51 & 4.3 & 43.3 & 2,881 & 3.0 & 50.6 \\
Codex \texttt{GPT-5.3} (high) & 42 & 27.1 & 343.5 & 2,494 & 2.6 & 373.2 \\
Codex \texttt{GPT-5.3} (xhigh) & 57 & 14.9 & 389.8 & 3,914 & 4.1 & 408.8 \\
\midrule
Total & 716 & \textbf{69.1} & 2845.9 & 38,099 & 44.5 & 2959.5 \\
\bottomrule
\end{tabular}%
}
\end{table}

\section{Descriptions of selected evolved models}
\label{app:model_descriptions}
\counterwithin{figure}{section}
\counterwithin{table}{section}
\renewcommand{\thefigure}{C\arabic{figure}}
\renewcommand{\thetable}{C\arabic{table}}

\cref{tab:agentic_imodels_library} summarizes the 10 evolved models highlighted in \cref{fig:overview}b.
Below, we describe in detail the three of these model classes that exemplify the two architectural patterns identified in \cref{sec:qualitative_model_descriptions}.
All models are \texttt{scikit-learn}-compatible and available in the released library.
To illustrate each model's \texttt{\_\_str\_\_} output concretely, we fit each on the same synthetic dataset:
$y = 2x_0 + 0.5x_1 + f(x_2) + \varepsilon$, where $f(x_2) = 1.5x_2$ for $x_2 > 0$ and $0.3x_2$ otherwise, $x_3$ is irrelevant, and $\varepsilon \sim \mathcal{N}(0, 0.3)$, with $n=300$.

\begin{table}[t]
\centering
\caption{The 10 evolved models highlighted in \cref{fig:overview}b.
\emph{Source} indicates which agent (Claude Code or Codex) discovered the model. \emph{Rank} is the normalized mean RMSE rank across the 65 development datasets in the global pool of evolved + baseline models in \cref{fig:overview}b (lower is better). \emph{Interp.} is the held-out interpretability score (fraction of 157 tests passed).}
\label{tab:agentic_imodels_library}
\footnotesize
\begin{tabular}{@{}l l c c p{6.1cm}@{}}
\toprule
Model & Source & Rank $\downarrow$ & Interp.\ $\uparrow$ & Description \\
\midrule
\texttt{HingeEBM} & Claude & 0.19 & 0.71 & LassoCV on a hinge (piecewise-linear) basis; hidden EBM corrects residuals (\cref{app:hinge_ebm}). \\
\texttt{DistilledTreeBlendAtlas} & Codex & 0.25 & 0.71 & Ridge student distilled from GBM/RF teachers with validation-calibrated nonnegative blending. \\
\texttt{DualPathSparseSymbolic} & Codex & 0.29 & 0.71 & Blended GBM/RF/Ridge teacher for batch prediction; sparse symbolic single-row equation for display. \\
\texttt{HybridGAM} & Claude & 0.29 & 0.68 & SmartAdditive display + hidden depth-5 random-forest residual, shrunk by $0.7$ (\cref{app:ridge_rf_resid}). \\
\texttt{TeacherStudentRuleSpline} & Codex & 0.36 & 0.80 & GBM teacher for batch path; sparse symbolic student over linear/spline/rule/gated terms for display. \\
\texttt{SparseSignedBasisPursuit} & Codex & 0.48 & 0.76 & Forward-selected signed basis (linear, hinge, square, interaction) with ridge refit and light coefficient rounding. \\
\texttt{HingeGAM} & Claude & 0.49 & 0.78 & Lasso on a hinge basis at 10 quantile knots, with adaptive SmartAdditive-style display. \\
\texttt{WinsorizedSparseOLS} & Claude & 0.58 & 0.73 & LassoCV-selected top-8 features after clipping inputs to the $[p_1, p_{99}]$ range; OLS refit. \\
\texttt{TinyDTDepth2} & Claude & 0.59 & 0.71 & Depth-2 (4-leaf) decision tree. \\
\texttt{SmartAdditive} & Claude & 0.62 & 0.73 & Greedy boosted-stump GAM with adaptive per-feature linearization (\cref{app:smooth_additive_gam}). \\
\bottomrule
\end{tabular}
\end{table}

\subsection{Describing the \texttt{HingeEBM\_5bag} model}
\label{app:hinge_ebm}

\paragraph{Architecture.}
\texttt{HingeEBM\_5bag} fits a two-stage model.
In stage 1, for each feature $j$, it constructs positive hinge terms $\max(0, x_j - t)$ and negative hinge terms $\max(0, t - x_j)$ at $K=2$ quantile knots, yielding a piecewise-linear basis expansion.
Lasso selects a sparse subset, producing the stage-1 prediction:
\begin{equation}
    \hat{y}^{(1)} = \beta_0 + \sum_j \beta_j x_j + \sum_{j,k} \left[\alpha_{jk}^+ \max(0,\, x_j - t_{jk}) + \alpha_{jk}^- \max(0,\, t_{jk} - x_j)\right].
    \label{eq:hinge_fit}
\end{equation}
In stage 2, if the residuals $y - \hat{y}^{(1)}$ explain more than 10\% of remaining variance, an EBM with 5 outer bags and 1{,}000 rounds is fitted on the residuals.
The final prediction is $\hat{y} = \hat{y}^{(1)} + \hat{y}^{(2)}_{\text{EBM}}$, but \texttt{\_\_str\_\_} displays only the stage-1 component.

\paragraph{Display logic.}
Rather than showing the hinge equation~\eqref{eq:hinge_fit} directly, \texttt{\_\_str\_\_} collapses each feature's hinge contributions into an effective linear slope.
For a positive hinge $\alpha^+ \max(0, x_j - t)$, the slope $\alpha^+$ is added to the coefficient of $x_j$ and $-\alpha^+ t$ is added to the intercept.
Negative hinges are handled symmetrically.
The result is a compact linear equation that is easy for an LLM to simulate, though it is only exact above the highest active knot.

\paragraph{Example output.}
\begin{verbatim}
Ridge Regression (L2 regularization, alpha=0.01043 chosen by CV):
  y = 1.9688*x0 + 0.5194*x1 + 1.5351*x2 + -0.1190

Coefficients:
  x0: 1.9688
  x2: 1.5351
  x1: 0.5194
  intercept: -0.1190
  Features with zero coefficients (excluded): x3
\end{verbatim}
The display correctly recovers $x_0$ and $x_1$ but shows a single slope for $x_2$ (1.54), which approximates the average effect of $x_2$'s piecewise-linear shape.
The hidden EBM captures the threshold nonlinearity.

\subsection{Describing the \texttt{SmoothAdditiveGAM} model}
\label{app:smooth_additive_gam}

\paragraph{Architecture.}
\texttt{SmoothAdditiveGAM} (\texttt{SmartAdditiveRegressor}, variant \texttt{smooth\_gam\_v1}) fits a greedy additive boosted-stump model.
In each of 200 rounds, it selects the feature $j$ and threshold $\tau$ that most reduces residual SSE via a depth-1 tree, updates the per-feature shape function $f_j$, and subtracts the fitted step.
After boosting, 3 passes of Laplacian smoothing (weights 0.6/0.2/0.2) are applied to each $f_j$.
The prediction is:
\begin{equation}
    \hat{y} = \mu + \sum_j f_j(x_j),
\end{equation}
where each $f_j$ is a piecewise-constant function over the thresholds chosen during boosting.

\paragraph{Display logic.}
For each feature $j$, a linear approximation $\hat{f}_j(x_j) \approx s_j x_j + b_j$ is computed on training data.
If $R^2 > 0.90$ and the feature contributes at least 1\% of total importance, it is rendered as a coefficient $s_j$; otherwise the full piecewise table is shown.
The Laplacian smoothing step is critical: it flattens irregular shape functions, increasing the fraction of features that pass the $R^2$ threshold and thus appear as simple linear terms.
(A later variant, \texttt{smooth\_gam\_r70hyb}, lowers this threshold to $R^2 > 0.70$ and adds a hybrid predict step; the core architecture is the same.)

\paragraph{Example output.}
\begin{verbatim}
Ridge Regression (L2 regularization, alpha=1.0000 chosen by CV):
  y = 1.8974*x0 + 0.3974*x1 + 0.5368

Coefficients:
  x0: 1.8974
  x1: 0.3974
  intercept: 0.5368

Nonlinear feature effects (piecewise corrections to add to above):
  f(x2):
    x2 <= -1.3747: -0.9201
    -1.3747 < x2 <= -0.9901: -0.7977
    -0.9901 < x2 <= -0.5740: -0.7116
    -0.5740 < x2 <= -0.3322: -0.6463
    ...
    x2 >  2.2444: +2.8050
\end{verbatim}
The display correctly identifies $x_0$ and $x_1$ as linear (recovering coefficients close to the ground truth of 2.0 and 0.5), and shows $x_2$ as a piecewise table that captures the slope change at 0.

\subsection{Describing the \texttt{RidgeRFResid} model}
\label{app:ridge_rf_resid}

To illustrate that display--predict decoupling was discovered independently across runs, we briefly describe \texttt{RidgeRFResid}.
Its architecture is analogous to \texttt{HingeEBM} but uses a different backbone and corrector:
\begin{equation}
    \hat{y} = \underbrace{\hat{y}_{\text{Ridge}}(x)}_{\text{displayed}} + \;\lambda \cdot \underbrace{\hat{y}_{\text{RF}}(x)}_{\text{hidden residual}},
\end{equation}
where $\hat{y}_{\text{Ridge}}$ is a Ridge regression model shown in \texttt{\_\_str\_\_} and a 100-tree depth-5 RandomForest fitted on residuals contributes to \texttt{predict} with shrinkage $\lambda = 0.6$.
This model achieved mean rank 6.6 with 67\% interpretability (within two rank positions of \texttt{HingeEBM\_5bag}) while displaying only a linear equation.
The same pattern re-emerged as the \texttt{HybridGAM} family in subsequent Claude runs, which used a SmartAdditiveGAM display instead of Ridge.

%%%%%%%%%%%%%%%%%%%%%%%%Add Checklist here at the end%%%%%%%%%%%%%%%%%%%%%%%%%%%%%%%%%%%%
\FloatBarrier

% % \newpage
% \input{checklist.tex}
\end{document}